\documentclass{article}
\usepackage[preprint]{neurips_2025}
\usepackage{placeins}
\bibpunct{[}{]}{;}{a}{,}{,}

\usepackage{amsmath}
\usepackage{amsthm}
\usepackage{amsfonts, amssymb, bbm, braket, accents}

\usepackage{graphicx}
\usepackage{wrapfig}
\usepackage[usenames,dvipsnames]{color}
\usepackage[makeroom]{cancel}
\usepackage{subcaption}
\usepackage{microtype}
\usepackage{tikz}
\usetikzlibrary{shapes.arrows}
\usetikzlibrary{shapes.geometric}
\usetikzlibrary{calc}
\usepackage{verbatim}
\usepackage{pifont}
\usepackage[mathscr]{euscript}
\usepackage{enumerate}
\usepackage{ragged2e}

\def\be{\begin{eqnarray}}
\def\ee{\end{eqnarray}}

\usepackage{multirow}
\usetikzlibrary{positioning}
\usepackage{caption}
\usepackage{enumitem}

\usepackage{booktabs}
\usepackage{xcolor}
\usepackage{siunitx}
\usepackage{colortbl}
\usepackage{algorithm,algpseudocode}

\newtheorem{theorem}{Theorem}

\theoremstyle{definition}

\usepackage{url}
\usepackage[breaklinks=true]{hyperref}

\usepackage[capitalise]{cleveref}
\crefformat{equation}{Eq.~(#2#1#3)}
\crefformat{section}{Sec.~#2#1#3}
\Crefformat{equation}{Equation~(#2#1#3)}
\crefformat{figure}{Fig.~#2#1#3}
\crefrangeformat{equation}{Eqs.~#3(#1)#4--#5(#2)#6}
\Crefformat{section}{Section~#2#1#3}

\title{Degree-Optimized Cumulative Polynomial Kolmogorov-Arnold Networks}

\author{%
  Mathew Vanherreweghe\\
  Centre for Quantum Technologies\\
  National University of Singapore\\
  Department of Mathematics\\
  National University of Singapore\\
  \texttt{v.mathew@u.nus.edu} \\
  \And
  Lirandë Pira \\
  Centre for Quantum Technologies\\
  National University of Singapore\\
  \texttt{lpira@nus.edu.sg} \\
  \And
  Patrick Rebentrost \\
  Centre for Quantum Technologies\\
  National University of Singapore\\
  Department of Computer Science\\
  National University of Singapore\\
  \texttt{cqtfpr@nus.edu.sg}
}

\begin{document}

\maketitle

\begin{abstract}
We introduce cumulative polynomial Kolmogorov-Arnold networks (CP-KAN), a neural architecture combining Chebyshev polynomial basis functions and quadratic unconstrained binary optimization (QUBO). Our primary contribution involves reformulating the degree selection problem as a QUBO task, reducing the complexity from $O(D^N)$ to a single optimization step per layer. This approach enables efficient degree selection across neurons while maintaining computational tractability. The architecture performs well in regression tasks with limited data, showing good robustness to input scales and natural regularization properties from its polynomial basis. Additionally, theoretical analysis establishes connections between CP-KAN's performance and properties of financial time series. Our empirical validation across multiple domains demonstrates competitive performance compared to several traditional architectures tested, especially in scenarios where data efficiency and numerical stability are important. Our implementation, including strategies for managing computational overhead in larger networks is available in Ref.~\citep{cpkan_implementation}.
\end{abstract}

\section{Introduction}\label{sec:introduction}
Many real-world problems involve complex, high-dimensional functions that are difficult or sometimes impossible to compute exactly. Approximating these functions allows one to make predictions, optimize systems, and analyze data in a computationally feasible way \cite{mhaskar2000fundamentals, trefethen2013approximation}. As such, function approximation plays a fundamental role in virtually any area of science including computational mathematics, machine learning, quantum computing and beyond. Efficient function approximation remains a central challenge in machine learning, particularly in domains requiring high data efficiency and numerical stability \citep{mitchell1997machine, bishop_pattern_2006, murphy2022machine}. While multilayer perceptrons (MLPs) are universal approximators \citep{Cybenko1989, krizhevsky2012imagenet, simonyan2014very, LeCun2015,  goodfellow_deep_2016, He2016}, their application can be hindered by sensitivity to input scaling (requiring normalization) and demands for large datasets \citep{Trihinas2024EvaluatingDM}. These limitations motivate exploring alternative architectures such as Kolmogorov-Arnold networks (KANs) \citep{liu_kan_2024}, inspired by the Kolmogorov-Arnold representation theorem \citep{kolmogorov_representation_1956}. Using learnable univariate activation functions on network edges, KANs aim for high expressiveness with potentially greater parameter efficiency than MLPs, showing an initial promise in scientific tasks. To achieve this expressiveness, various KAN architectures have explored different bases, including splines \citep{liu_kan_2024}, wavelets \citep{Bozorgasl2024}, and Chebyshev polynomials \citep{ss2024chebyshevpolynomialbasedkolmogorovarnoldnetworks}. While theoretical work has explored quantum implementations of Chebyshev KANs \citep{ivashkov2024qkanquantumkolmogorovarnoldnetworks}, practical classical implementations may face challenges, notably in degree selection.

A crucial challenge with polynomial bases in KANs is selecting the appropriate polynomial degree for each activation function.\citep{ polykans} The degree choice significantly impacts model expressivity, computational cost, and stability\citep{Ivakhnenko1970PolynomialTO, Sierra2001,park2004self}.Low degrees limit power while high degrees risk overfitting and increase computational overhead.\citep{ss2024chebykan} Different activations may need different degrees, making manual selection or simple heuristics suboptimal. This degree selection is a discrete optimization problem, difficult to integrate into gradient-based training and computationally prohibitive ($O(D^N)$ complexity for $N$ activations and maximum degree $D$) for exhaustive search \citep{carpenter1992artmap, chebyshev_encrypted_cnn, quantum_chebyshev_network}.

To address this challenge, we introduce cumulative polynomial Kolmogorov-Arnold networks (CP-KAN). Our approach uses Chebyshev polynomials and tackles degree selection via quadratic unconstrained binary optimization (QUBO), suitable for discrete optimization \citep{Yarkoni2022quantum}. We reformulate selecting the optimal degree for each neuron's polynomial activation as a QUBO problem. This allows efficiently finding optimal degree assignments for all neurons in a layer simultaneously using simulated annealing \citep{kirkpatrick1983optimization, neal2020,mo2021sa-nas}. This layer-wise QUBO formulation substantially reduces the complexity from the intractable $O(D^N)$ search to a single, feasible optimization step per layer, enabling adaptive structure optimization within the KAN framework\citep{chen2023efficient}. The resulting CP-KAN layer architecture is shown in \Cref{fig:cpkan-layer}.

Following background in \Cref{sec:background}, \Cref{sec:architecture} presents the CP-KAN architecture, detailing its integration of Chebyshev polynomials and our QUBO-based strategy for adaptive degree selection, reducing optimization complexity. This section also connects CP-KAN to financial time series properties. \Cref{sec:numerical} provides empirical results, demonstrating the model's performance and parameter efficiency, particularly in regression. \Cref{sec:discussion} analyzes these results, discusses implications and limitations, and suggests future work. Finally, \Cref{sec:conclusion} summarizes our contributions.

\section{Background and Methods}\label{sec:background}
\subsection{Kolmogorov-Arnold Networks}
Kolmogorov-Arnold networks \citep{liu_kan_2024} represent a promising architecture for efficient machine learning, building on the foundational Kolmogorov-Arnold representation theorem \citep{arnold_functions_1957, arnold_representation_1959}. This theorem establishes that any continuous multivariate function can be represented through compositions of univariate functions. Recent work has shown how quantum resources can enhance KAN implementations through quantum annealing optimization \citep{Yarkoni2022quantum}, particularly when using appropriate basis functions. Specifically, the Kolmogorov-Arnold theorem proves that any continuous function $f: [0,1]^n \rightarrow \mathbb{R}$ can be expressed as:
\begin{align}
f(x_1,\ldots,x_n) = \sum_{q=0}^{2n} \Phi_q\left(\sum_{p=1}^n \psi_{p,q}(x_p)\right)
\end{align}
where $\Phi_q$ and $\psi_{p,q}$ are continuous univariate functions \citep{kolmogorov_representation_1956}. This representation naturally suggests a neural architecture where inner and outer functions are learned to approximate complex mappings \citep{liu_kan_2024}. Unlike traditional neural networks that may require significant parameter tuning and large datasets, KANs utilize the Kolmogorov-Arnold theorem to represent complex functions with fewer parameters. While early implementations used b-splines and wavelets \citep{Bozorgasl2024}, we focus on Chebyshev polynomials for their orthogonality and efficient spectral convergence for smooth functions.\cite{chebyshev_spectral_method}

\subsection{Chebyshev Polynomials and Degree Optimization}
The Chebyshev polynomials of the first kind, denoted as $T_n(x)$, form a complete orthogonal basis on $[-1,1]$ and are defined by the recurrence relation:
\begin{align}
T_0(x) &= 1 \\
T_1(x) &= x \\
T_{n+1}(x) &= 2xT_n(x) - T_{n-1}(x)
\end{align}
While previous work such as QKAN \citep{ivashkov2024qkanquantumkolmogorovarnoldnetworks} first introduced these polynomials to KAN architectures and explored their quantum implementations, scaling and degree selection remained significant challenges. Our approach makes use of these polynomials in a classical framework optimized for practical applications.

Recent work in efficient architecture optimization \citep{chen2023efficient, Sierra2001,park2004self} has highlighted the importance of discrete optimization in neural network design. Building on these insights, our primary contribution involves reformulating the polynomial degree selection problem through QUBO optimization, enabling efficient classical implementation while maintaining the theoretical advantages of Chebyshev bases. In our framework, each activation function is defined as:
\begin{align}
\varphi_{pq}(x) = \frac{1}{d+1} \sum_{r=0}^d w_{pq}^{(r)} T_r(x)
\end{align}
where $w_{pq}^{(r)} \in [-1,1]$ are trainable weights, $d$ is the maximum polynomial degree, and the specific degrees of $T_r(x)$ up to $d$ are selected through our QUBO formulation. Unlike approaches that rely solely on gradient-based optimization \citep{liu2023gradientfree}, our classical QUBO formulation enables efficient degree selection while optimizing polynomial combinations across neurons\citep{chen2023efficient}. The approach enforces regularization through bounded coefficients, building on advances in binary optimization for neural architecture search \citep{zhang2024binary}, applying these techniques specifically to polynomial degree selection and potentially positioning them for future quantum-based optimizations through annealing \citep{Yarkoni2022quantum} or leveraging Grover-style quantum speedups \citep{Gilliam2021groveradaptive}. Our implementation can provide practical utility on current hardware. The cumulative polynomial structure further enables representation of hierarchical features, potentially beneficial for financial time series analysis \citep{dahlhaus2012locally}.

\section{CP-KAN Architecture and Implementation}\label{sec:architecture}

\subsection{Network Architecture}
\begin{figure}[t]
    \centering
    \begin{tikzpicture}[
        scale=0.8,
        transform shape,
        node distance=1.8cm,
        neuron/.style={circle,draw=black,minimum size=0.7cm},
        poly/.style={rectangle,draw=black,minimum width=1.1cm,minimum height=0.7cm},
        combine/.style={circle,draw=black,fill=gray!20,minimum size=0.7cm},
        >=latex
    ]

        \foreach \y/\name in {0/x_1,1/x_2,2/\vdots,3/x_n} {
            \node[anchor=east] (x\y) at (-4,-\y) {$\name$};
        }
        
        \foreach \y/\d in {0/d_1,1/d_2,2/\vdots,3/d_k} {
            \node[neuron] (p\y) at (-2,-\y) {$\alpha$};
            \node[poly] (t\y) at (0,-\y) {$T_{\d}$};
            \draw[->] (p\y) -- (t\y) node[midway,above] {\tiny proj};
        }
        
        \foreach \x in {0,1,2,3} {
            \foreach \y in {0,1,2,3} {
                \draw[->] (x\x) -- (p\y);
            }
        }
        
        \node[combine] (out) at (3,-1.5) {$\sum$};
        
        \foreach \y in {0,1,2,3} {
            \draw[->] (t\y) -- (out);
        }
        
        \node[anchor=west] (y) at (5,-1.5) {$y$};
        \draw[->] (out) -- (y);
        
        \node[align=center] at (-2,-4.5) {Projection\\Neurons};
        \node[align=center] at (0,-4.5) {Chebyshev\\Polynomials};
        \node[align=center] at (3,-4.5) {Linear\\Combination};
    \end{tikzpicture}
    \caption{Single layer architecture of CP-KAN showing input features ($x_1,\ldots,x_n$), 
    projection neurons ($\alpha$), Chebyshev polynomial transformations ($T_{d_i}$), and 
    their linear combination for output prediction. Each projection neuron learns optimal 
    weights and bias for its input transformation.}
    \label{fig:cpkan-layer}
\end{figure}
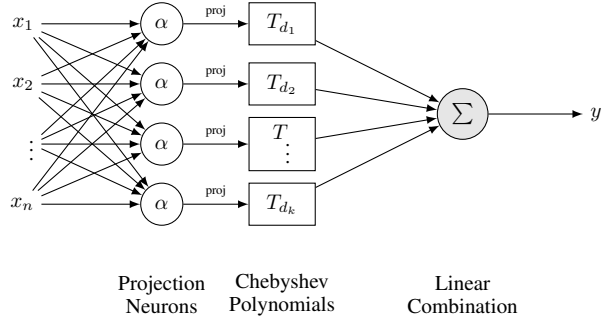

The CP-KAN framework builds upon the Chebyshev-based Kolmogorov-Arnold framework. At its core lies the KANNeuron, characterized by the triplet $(\mathbf{w},b,d)$ where $\mathbf{w}\in\mathbb{R}^n$ are projection weights, $b\in\mathbb{R}$ is the projection bias and $d\in\mathbb{N}$ is the polynomial degree which is either QUBO-optimized or fixed. Each neuron implements the following transformation:
\begin{align}
f(\mathbf{x}) 
= \sum_{i=0}^d \,c_i\, T_i(\alpha(\mathbf{x})), 
\quad 
\alpha(\mathbf{x}) = \mathbf{w}^\top\mathbf{x} + b,
\end{align}
where $T_i$ denotes the Chebyshev polynomials of the first kind. A CP-KAN layer maps $n_{\mathrm{in}}$ inputs to $n_{\mathrm{out}}$ outputs through parallel scalar-output neurons, as in \Cref{fig:cpkan-layer}. The layer combines these outputs through a learned linear transform, enabling feature interactions. Multiple such layers can be stacked to form a deep CP-KAN architecture.

\subsection{Degree Selection Optimization}
\label{sec:qubo}
The challenge of selecting optimal polynomial degrees $d_i$ for each neuron $i$ is a discrete optimization problem, difficult to integrate into gradient descent and computationally prohibitive for exhaustive search. This degree assignment can be formulated as a 0-1 integer program with a one-hot constraint per neuron. For compatibility with quantum and annealing solvers \citep{Yarkoni2022quantum}, we express it as a QUBO by adding a penalty term for the one-hot constraint. Using binary variables $q_{(i,d)}$ (where $q_{(i,d)}=1$ when neuron $i$ uses degree $d$), we create the objective:
\begin{align}
\min_{\{q_{(i,d)}\}\in\{0,1\}} 
\;\;\Bigl(\sum_{i,d} \text{MSE}_{i,d} \cdot q_{(i,d)}
\;+\; 
\alpha\sum_i(\sum_d q_{(i,d)}-1)^2\Bigr).
\label{eq:qubo_objective}
\end{align}
The first term aims for accurate function approximation through partial least-squares, while the second term enforces a single degree assignment per neuron. A QUBO solver yields optimal degree selections $\mathbf{q}^*$, after which we determine the polynomial coefficients $c_i$. See \Cref{sec:appendix_qubo_details} for the complete formulation.

\subsection{Training Strategy and Two-Phase Optimization}
Our training process employs a two-phase strategy combining discrete structure optimization with continuous parameter refinement, designed for hardware agnosticism and potential future quantum readiness\citep{noy2020asap,e24030348}. In Phase 1 (Degree Optimization), the QUBO formulation (\Cref{eq:qubo_objective}) is solved layer-by-layer using simulated annealing to determine the optimal polynomial degree $d_i$ for each neuron $i$. This phase identifies an effective degree structure. As a practical consideration for large layers where the QUBO problem size ($N \times (D+1)$) exceeds a threshold $T$, we may skip QUBO and default to a fixed degree (i.e., $d=4$) to manage computation, see \Cref{alg:cpkan-training}. In Phase 2 (Parameter Optimization), following degree selection, all continuous trainable parameters—projection weights $\mathbf{w}$, biases $b$, and optionally the polynomial coefficients $\mathbf{c}$—are fine-tuned using standard gradient-based optimization (i.e., Adam \citep{kingma2015adam}) with an appropriate loss function (i.e., MSE or Cross-Entropy) over a set number of epochs. The detailed pseudocode for this process is presented in \Cref{alg:cpkan-training} in \Cref{sec:appendix_training_algorithm}.

\subsection{Implementation Considerations}
While our QUBO formulation provides a structured approach to degree selection, practical implementation requires careful consideration of computational resources. Memory requirements scale with network size and maximum polynomial degree. Each neuron-degree pair requires storing partial MSE results, creating a memory footprint of shape $(\text{neurons}) \times (\text{max\_degree}+1)$. Solver complexity is tied to the dimension $M$ of the QUBO problem. Our empirical analysis confirms these scaling relationships, with optimization time scaling approximately linearly $O(N)$ with network size $N$ (see \Cref{fig:QUBO_log_scaling}). While other optimization methods were explored (see \Cref{sec:appendix_optim_formulations}), QUBO offered a good balance of performance consistency across domains and computational efficiency, particularly for larger networks. In our implementations, we found problems up to $\sim$5000 binary variables ($N \times (D+1)$) feasible on typical CPU/GPU clusters. Beyond this, alternative strategies such as fixing polynomial degrees for some layers or using the QUBO skipping mechanism (\Cref{alg:cpkan-training}) provide a balance between optimization quality and computational efficiency.

The effectiveness of CP-KAN, particularly for financial time series, is grounded in its ability to capture locally stationary processes. Chebyshev polynomial basis functions ($T_i$) are efficient for representing the smooth, time-varying spectral characteristics ($A_t(\omega)$) inherent in these processes (see \Cref{sec:appendix_mean_reversion_theory} for details on coefficient decay \citep{decayRateChebyshev}). The CP-KAN framework, using QUBO (\cref{sec:qubo}) for adaptive degree selection, gains flexibility to model the complex, evolving dynamics characteristic of local stationarity \citep{dahlhaus2012locally}, learning underlying patterns while adapting to local variations.

\section{Experiments}\label{sec:numerical}
\subsection{Datasets Overview}
Our analysis utilizes several datasets to evaluate CP-KAN across different tasks. For regression, we use the high-dimensional Jane Street Market Prediction dataset \citep{janestreet2024} and the standard House Price prediction task \citep{Grinsztajn2022tabular}. For classification, we use MNIST, CIFAR-10, and the Forest Covertype dataset \citep{Grinsztajn2022tabular}. Full details on data sources, preprocessing, splits, and configurations for each dataset are provided in \Cref{sec:appendix_data_config} and the respective experimental Appendices \ref{sec:appendix_training_algorithm}, \ref{sec:appendix_jane_street_hyperparams}, \ref{sec:appendix_house_sales_hyperparams}.

\subsection{Model Architectures and Parameter Control}
To ensure fair comparisons, all evaluated neural network models (CP-KAN, MLPs) were configured to have approximately the same number of trainable parameters ($\sim$10,000), controlling for model capacity, unless otherwise stated. Specific architectural choices (i.e., hidden layers, activation functions) and the hyperparameter search spaces used for each dataset are detailed in the corresponding appendices: Jane Street (\Cref{sec:appendix_jane_street_hyperparams}), House Price (\Cref{sec:appendix_house_sales_hyperparams}), MNIST/CIFAR-10 (\Cref{sec:appendix_optim_eval_mnist_cifar}), and Covertype (\Cref{sec:appendix_covertype_exp_config}). Baseline models such as RandomForest and XGBoost used standard configurations (\Cref{tab:covertype_baseline_configs}). All models were implemented in PyTorch \citep{pytorch2019} and trained with the Adam optimizer unless noted otherwise.

\subsection{Feature Reshaping for Sequential Models}
To adapt sequential models (LSTM, GRU, Transformer) for tabular datasets similar to Jane Street, we employed a feature reshaping technique, treating each feature as an individual step or token in a sequence. This allows standard sequence architectures to process the data. Full implementation details of this embedding trick and the specific recurrent/transformer architectures used are described in \Cref{sec:appendix_recurrent_transformer}.

\subsection{Financial Market Prediction}\label{subsec:finmar}

\begin{figure}[t]
    \centering
    \includegraphics[width=0.99\textwidth]{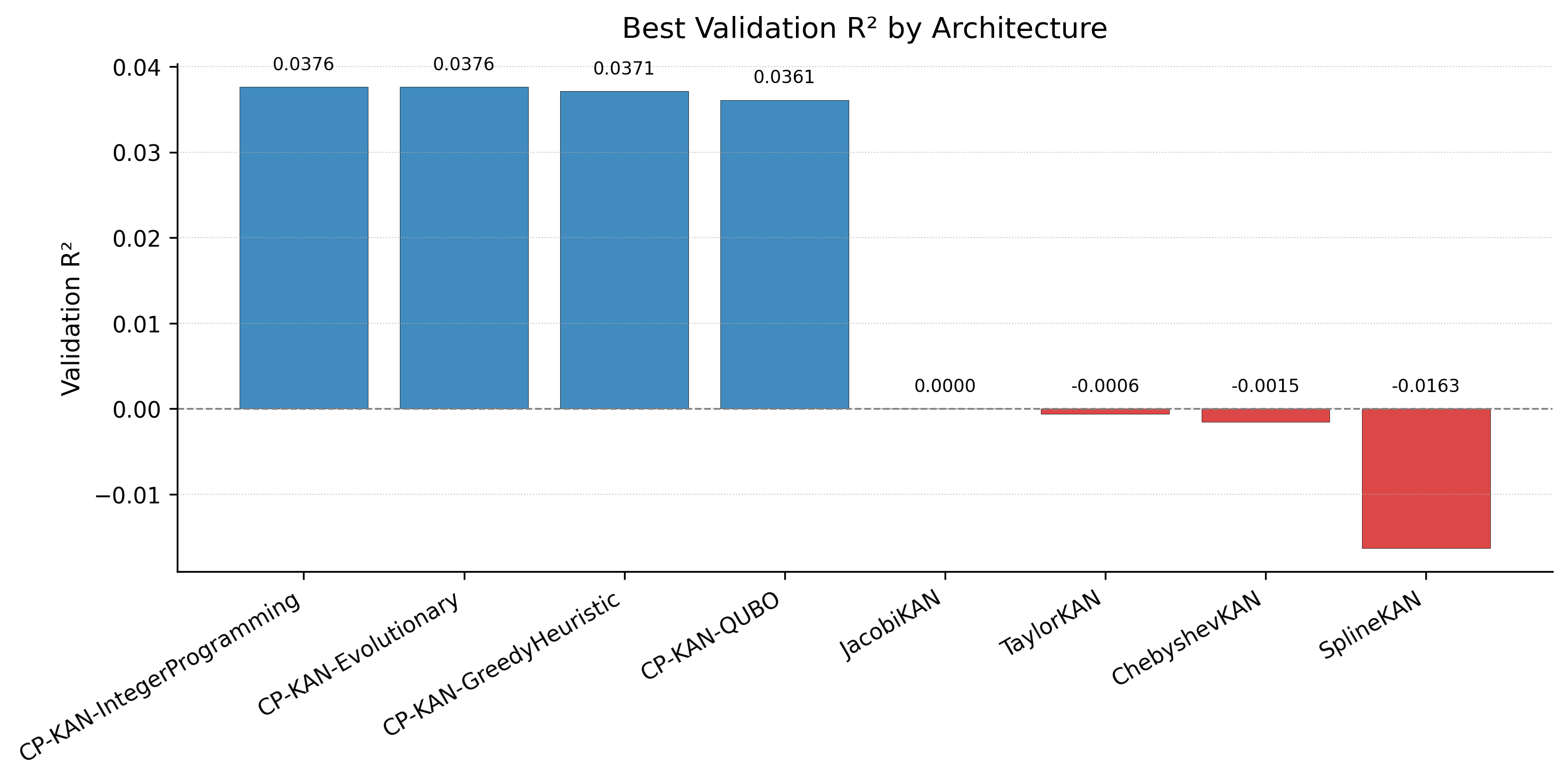}
    \caption{Performance comparison across KAN architectures on the Jane Street Market Prediction dataset. All CP-KAN variants achieve stronger $R^2$ values than other KAN architectures.}
    \label{fig:kan_comparison}
\end{figure}

\begin{figure}[t]
  \centering
  \begin{subfigure}[t]{0.49\textwidth}
    \centering
    \includegraphics[width=\linewidth]{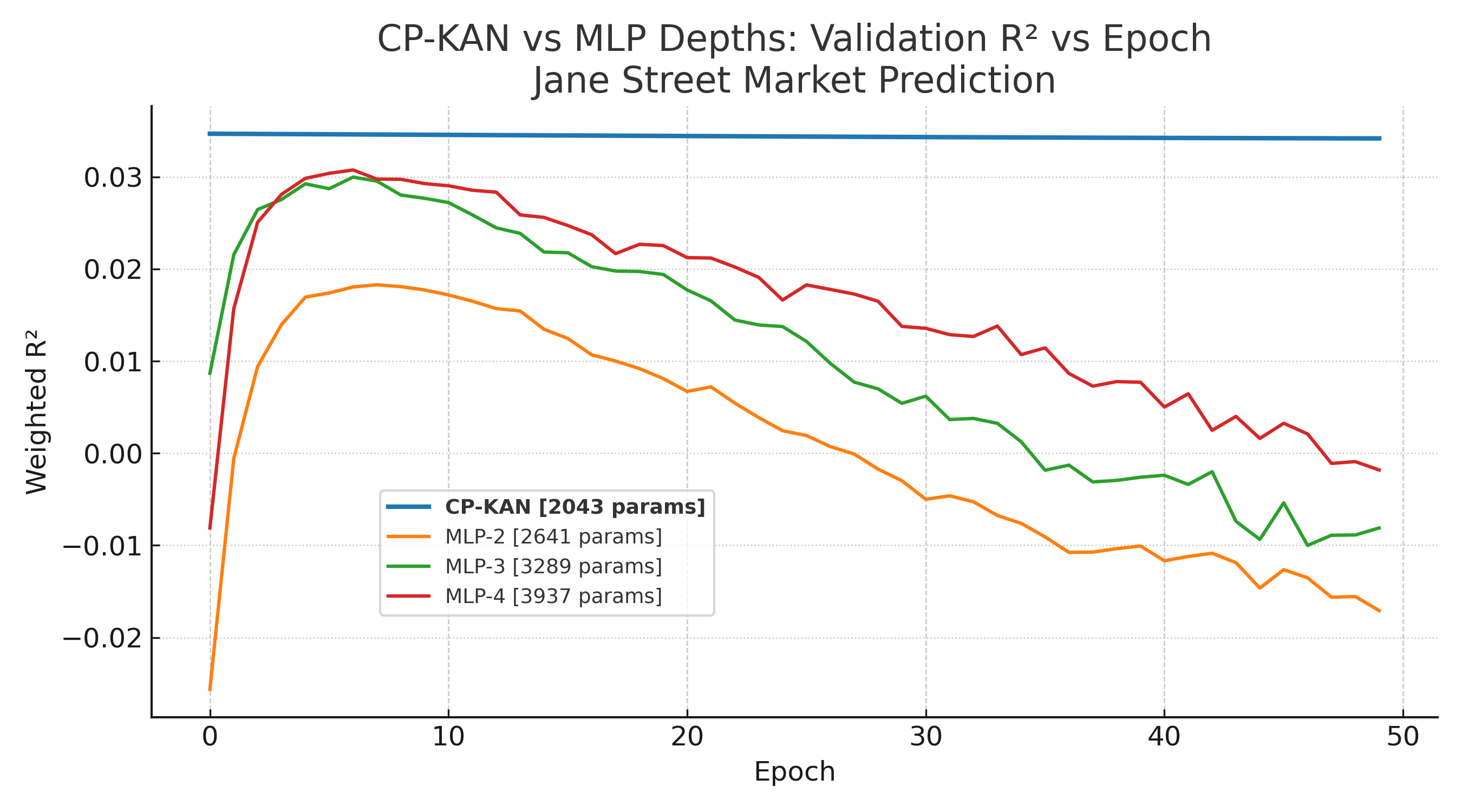}
    \caption{Jane Street dataset: CP-KAN vs.\ MLP validation performance.}
    \label{fig:training_curves}
  \end{subfigure}
  \hfill
  \begin{subfigure}[t]{0.49\textwidth}
    \centering
    \includegraphics[width=\linewidth]{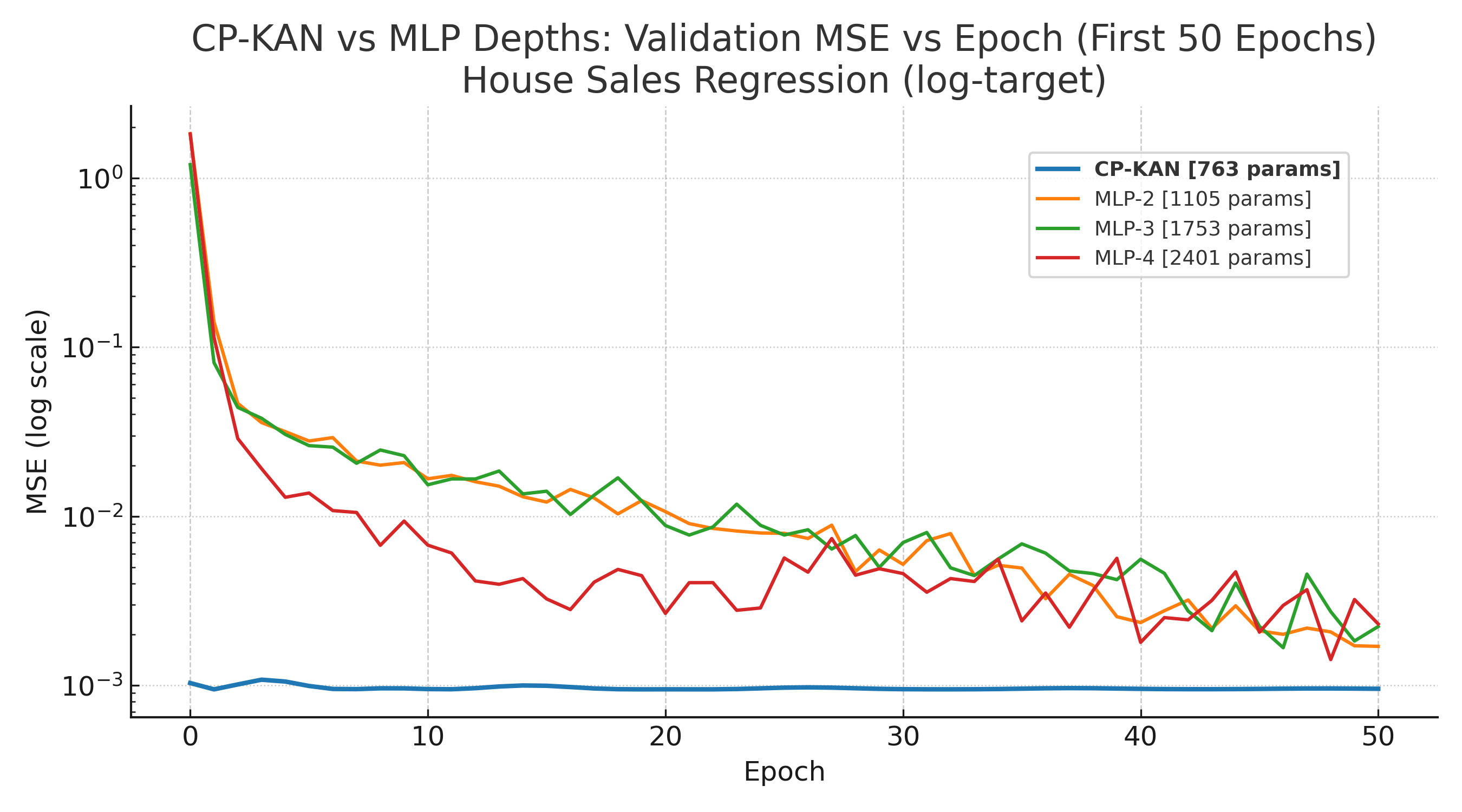}
    \caption{House price prediction: CP-KAN vs.\ MLP training dynamics.}
    \label{fig:house_training}
  \end{subfigure}
  \caption{Comparison of training behavior and generalization between CP-KAN and MLPs across two regression tasks. CP-KAN shows more stable and consistent performance.}
  \label{fig:kan_vs_mlp_comparison}
\end{figure}

The Jane Street Market Prediction task provides a rigorous test of model generalization in a real-world financial context  with a custom evaluation metric, defined as: 
\begin{align}
    \text{R}^2 = 1 - \frac{\sum w_i(y_i - \hat{y}_i)^2}{\sum w_i y_i^2}.
    \label{eq:weighted_r2}
\end{align}
where $y$ and $\hat{y}$ denote the vectors of ground-truth and predicted values for the response variable under evaluation, and $w$ is the non-negative sample-weight vector. 
Our experiments focus on a specific trading window to evaluate performance under limited data conditions. We note that similar results were observed on other trading days with similar data densities. Illustrative training curves showing CP-KAN's stable convergence compared to MLP on this task can be found here \Cref{fig:training_curves}. To contextualize CP-KAN's performance, we conducted extensive comparisons against state-of-the-art deep learning architectures widely used for financial time series prediction. We implemented and optimized LSTM, Transformer, and competition-scale GRU models on the Jane Street Market Prediction dataset, as in \Cref{sec:appendix_recurrent_transformer}. As shown in \Cref{tab:kan_performance}, both CP-KAN variants achieve competitive performance compared to all deep learning models while using orders of magnitude fewer parameters. These $R^2$ scores, while seemingly low, are highly competitive for this challenging financial prediction task where even small improvements in predictive power can translate to significant trading advantages. Despite employing the feature reshaping technique to optimize performance of sequence-based architectures, CP-KAN still achieved competitive results with fewer parameters. As shown in \Cref{tab:kan_performance}, transformers and LSTMs yielded respectable performance but were consistently outperformed by CP-KAN variants with fewer parameters. 

\begin{figure}[t]
    \centering
    \begin{subfigure}[t]{0.49\textwidth}
        \centering
        \includegraphics[width=\textwidth]{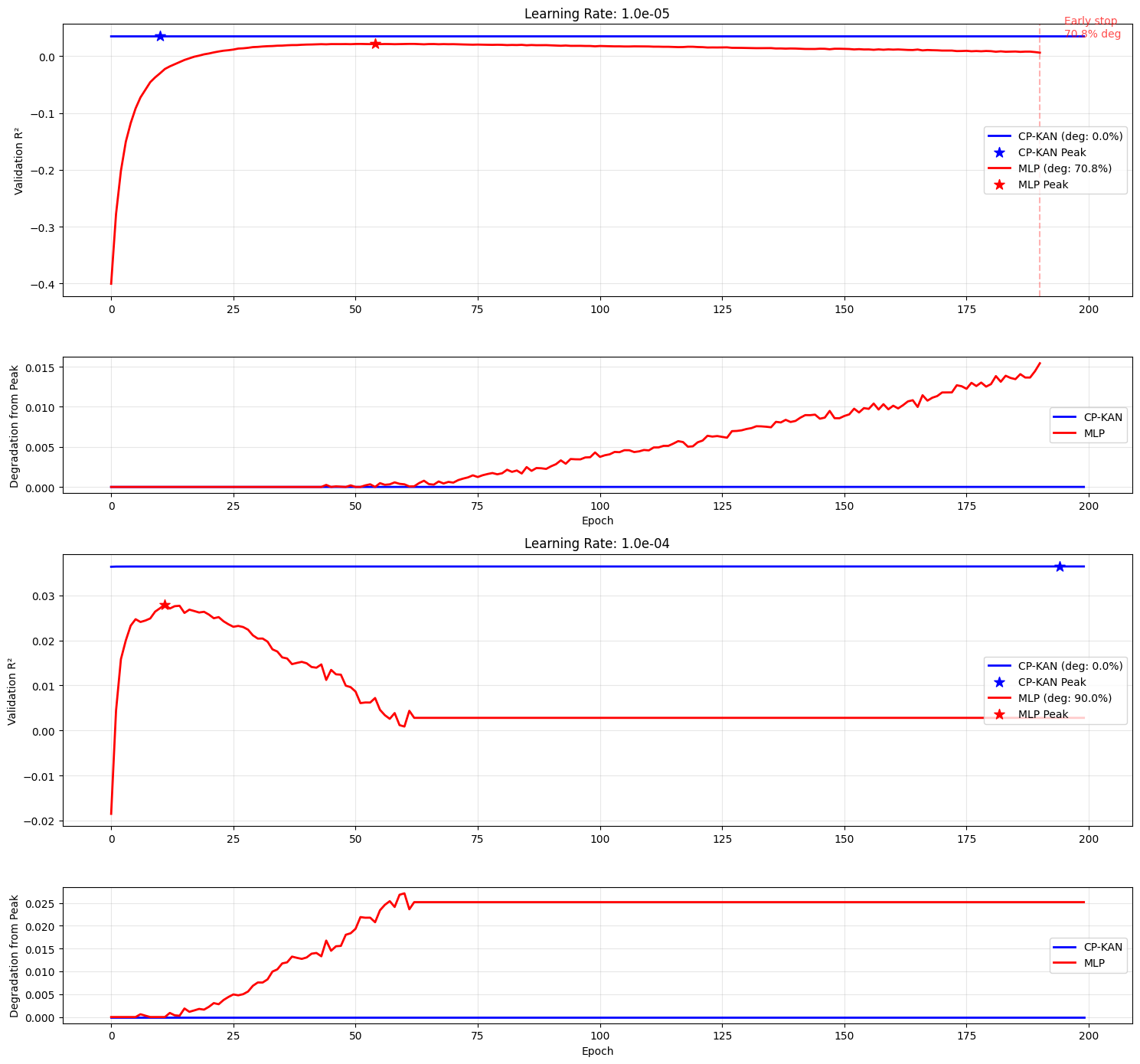}
        \caption{Lower learning rates ($1e\!-\!5$, $1e\!-\!4$).}
        \label{fig:stability_low_lr}
    \end{subfigure}
    \hfill
    \begin{subfigure}[t]{0.49\textwidth}
        \centering
        \includegraphics[width=\textwidth]{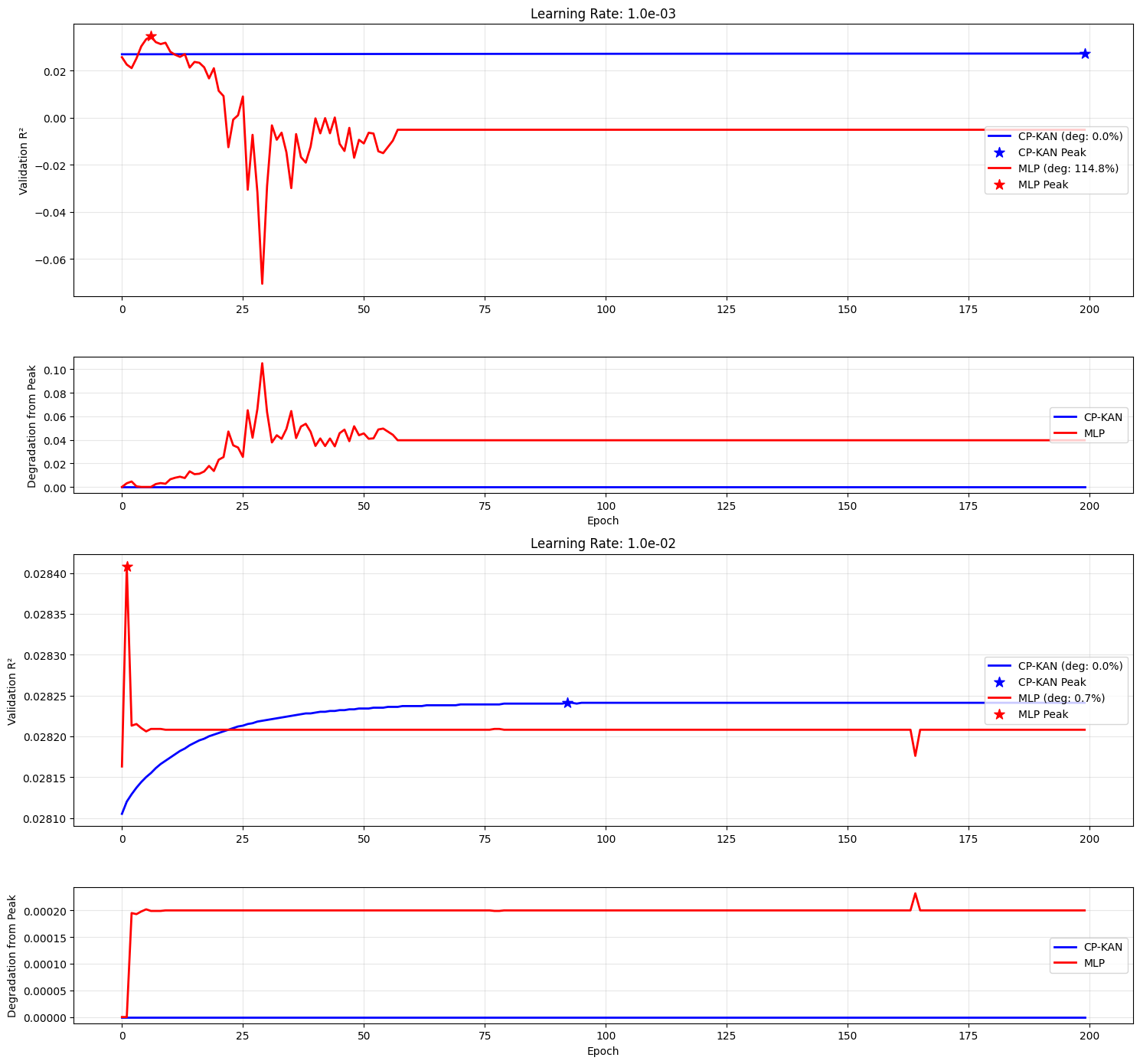}
        \caption{Higher learning rates ($1e\!-\!3$, $1e\!-\!2$).}
        \label{fig:stability_high_lr}
    \end{subfigure}
    \caption{Training stability comparison between MLP and CP-KAN on the Jane Street Market Prediction dataset across varying learning rates.}
    \label{fig:stability_comparison}
\end{figure}

Our comparative analysis against other KAN architectures as seen in \Cref{fig:kan_comparison} further validates our approach, with CP-KAN using different optimization methods such as integer programming, greedy heuristic and evolutionary, and $1500-2500$ parameters achieved strong results while other polynomial-based KANs fail to exceed mean predictor performance using $6000-9500$ parameters after a thorough hyperparameter search. Notably, the standard Chebyshev KAN without our optimization strategy obtained a negative $R^2$ score, highlighting the importance of the QUBO-based degree selection \citep{han2023are}. Additionally, ablation studies through hyperparameter grid search finding the best hyperparameters revealed a crucial stability advantage of CP-KAN over traditional architectures for this financial task. As shown in \Cref{fig:stability_comparison}, when the best MLP and KAN models identified by the grid search were trained for additional epochs, MLPs initially achieved competitive performance but exhibited significant performance degradation over extended training periods across multiple learning rates ($1e-5$ to $1e-2$). In contrast, CP-KAN maintained stable and consistent performance throughout, with minimal deviation from peak performance.

\subsection{House Price Regression}
Our house price prediction experiments suggest CP-KAN's robustness to input scales, an advantage in real-world regression tasks. The results in \Cref{tab:combined_results} illustrate this robustness. CP-KAN outperforms MLPs in terms of mean squared errors (MSE) when dealing with raw house prices, indicating stability advantages. This robustness suggests CP-KAN's potential for handling certain real-world regression tasks without explicit preprocessing. The training dynamic comparison can be found in \Cref{fig:house_training}).

\begin{table*}[ht]
    \centering
    \renewcommand{\arraystretch}{1.2}
    \setlength{\tabcolsep}{3pt}
    \small
    \caption{Performance comparison across two tasks: Jane Street market prediction dataset (Validation $R^2$) and house price prediction dataset (mean squared errors), with observations aligned per task.}
    \label{tab:combined_results}
    \begin{tabular}{l | p{2cm} p{3cm} | p{1.5cm} p{1.5cm} p{3cm}}
        \toprule
        \textbf{Model} & \multicolumn{2}{c|}{\textbf{Market Prediction: Validation $R^2$}} & \multicolumn{3}{c}{\textbf{House Price: Mean Squared Errors (MSE)}} \\
        \cmidrule(lr){2-3} \cmidrule(lr){4-6}
        & Validation $R^2$ & Observation & Raw MSE & Log MSE & Observation \\
        \midrule
        CP-KAN (ours) & 0.0376 & Consistent train-validation performance & 0.29 & 0.0009 & Robust to scaling \\
        MLP-4 & 0.0296 & Strong initial performance but unstable, degrades to -0.0042 & 1823.34 & 0.0009 & Requires normalization \\
        Best Tree Model & 0.0293 & Strong baseline performance, balancing bias and variance & - & - & - \\
        Random Forest & - & - & 0.03 & 0.0002 & Best overall \\
        XGBoost & - & - & 0.03 & 0.0002 & Ties RF \\
        \bottomrule
    \end{tabular}
\end{table*}

\begin{table*}[t]
    \centering
    \renewcommand{\arraystretch}{1.2}
    \setlength{\tabcolsep}{5pt}
    \small
    \caption{Detailed $R^2$ performance statistics across model architectures on Jane Street (\ref{subsec:finmar}). CP-KAN variants achieve superior performance with dramatically fewer parameters ($1,571-2,547$) compared to traditional deep learning models ($17,185-506,701$), while significantly outperforming other KAN architectures ($6,693-25,313$) despite their larger parameter counts.}\label{tab:kan_performance}
    \begin{tabular}{lrrr}
        \toprule
        \textbf{Architecture} & \textbf{Mean R²} & \textbf{Min R²} & \textbf{Max R²} \\
        \midrule
        CP-KAN-IntegerProgramming (2,547 params) & 0.0188 & 0.0000 & 0.0376 \\
        CP-KAN-Evolutionary (1,571 params) & 0.0188 & 0.0000 & 0.0376 \\
        CP-KAN-GreedyHeuristic (1,571 params) & 0.0186 & 0.0000 & 0.0371 \\
        CP-KAN-QUBO (1,571 params) & 0.0241 & 0.0122 & 0.0361 \\
        \midrule
        Transformer (17,185 params) & 0.0345 & 0.0297 & 0.0363 \\
        LSTM (37,185 params) & 0.0312 & 0.0132 & 0.0375 \\
        GRU Large (506,701 params) & 0.0320 & 0.0144 & 0.0361 \\
        \midrule
        ChebyshevKAN (7,609 params) & -0.4644 & -4.6551 & -0.0015 \\
        JacobiKAN (9,505 params) & -0.6273 & -4.0507 & -0.0000 \\
        SplineKAN (25,313 params) & -0.6985 & -4.1313 & -0.0163 \\
        TaylorKAN (6,693 params) & -0.3540 & -3.7172 & -0.0006 \\
        \bottomrule
    \end{tabular}
\end{table*}

\section{Experimental Results and Discussion}\label{sec:discussion}
\subsection{Domain-Specific Behavior}
Our numerical evaluation indicates CP-KAN's performance in regression tasks, particularly with limited data and when scale robustness is important. The results suggest stability advantages over traditional neural architectures, supported by its performance on raw house prices without normalization and its resistance to performance degradation in financial time series prediction.

Our comparative analysis offers potential insights into why CP-KAN outperforms traditional deep learning architectures on the tested financial datasets. While our feature reshaping approach (\Cref{sec:appendix_recurrent_transformer}) enabled sequence models to process tabular data, these architectures still struggled with generalization. This might relate to optimization difficulties potentially exacerbated by limited data; sequential models may struggle to learn generalizable patterns from fewer examples or converge to suboptimal solutions when processing arbitrarily ordered features \citep{Trihinas2024EvaluatingDM,mcelfresh2024neuralnetsoutperformboosted, ren2025deeplearningtabulardata}. The inductive bias of these models assumes sequential relationships, which may not align with the structure of the financial feature vectors tested. In contrast, CP-KAN's design may be well-suited to the data's underlying statistical properties, potentially enabling generalization with fewer parameters.

This domain-dependent behavior suggests a relationship between data characteristics and model stability. In financial time series with high temporal dependency and non-stationarity, MLPs exhibited performance degradation after extended training (\Cref{fig:stability_low_lr}, \Cref{fig:stability_high_lr}), while CP-KAN appeared to maintain consistent performance. This stability gap was less pronounced in housing price prediction (\Cref{fig:house_training}). The polynomial basis may provide regularization benefits in domains with higher temporal dependency, where traditional architectures may be more prone to overfit transient patterns.

The optimization methods within our CP-KAN framework also exhibit domain-dependent trade-offs. While Integer Programming achieved marginally higher peak performance on financial data, QUBO appeared robust across diverse tasks~\citep{chen2023efficient}. This cross-domain stability may relate to QUBO's ability to navigate the discrete optimization landscape while being potentially less susceptible to local minima compared to greedy approaches. 

\subsection{Parameter Efficiency}
CP-KAN resulted in competitive performance with considerably fewer parameters compared to the tested traditional deep learning architectures and other KAN variants (\Cref{tab:kan_performance} and \Cref{fig:kan_comparison}). This efficiency might relate to two key factors: First, the Chebyshev polynomial basis provides an inductive bias potentially suitable for the financial time series tested, capturing underlying dynamics effectively. Second, our QUBO-based degree optimization attempts to match polynomial complexity appropriate for its specific function, avoiding over-parameterization\citep{Sierra2001, park2004self}.

The performance of CP-KAN in financial time series prompts consideration of why the sequence models, despite their expressive power, did not match CP-KAN's performance with fewer parameters on this task. One potential explanation is that CP-KAN's single-feature polynomial transformations may be better suited for the financial data characteristics than the multi-feature learned transformations in other architectures. While sequential models attempt to learn complex dependencies between features treated as sequence steps, they may lack the appropriate inductive bias for the statistical properties of the financial data tested. Additionally, the optimization landscape for such models may be more complex, potentially leading to convergence issues that CP-KAN's approach potentially avoids.

While the ``bigger is better'' paradigm drives much of deep learning, this efficiency underscores the continued relevance of tailored architectures with suitable inductive biases for specific domains. CP-KAN's performance hints that carefully designed, domain-appropriate architectures might outperform generic scaled-up models in certain scenarios. The restricted polynomial degree provides implicit regularization, enhancing parameter efficiency and reducing overfitting risk, especially with limited data. Additionally, the polynomial basis may yield a smoother optimization landscape compared to deeper networks, potentially aiding generalization.

\subsection{Theoretical Justification}
CP-KAN's effectiveness for financial time series can be theoretically linked to its ability to capture mean-reverting processes through Chebyshev polynomials. The infinitesimal generator for processes such as the Ornstein-Uhlenbeck SDE admits a Chebyshev expansion (see \Cref{sec:appendix_mean_reversion_theory} for the formal theorem \citep{infinitesimalGen2007}), offering a potential justification for using these polynomials to model financial dynamics. Theoretical error bounds highlight two aspects: an exponential decay term capturing mean reversion, and a term related to polynomial degree showing diminishing returns for higher degrees (\Cref{sec:appendix_mean_reversion_theory} \citep{spectralPerturb2018}). This mathematical foundation helps explain why CP-KAN might capture underlying financial patterns where models lacking this inductive bias might struggle.

\subsection{Scaling and Classification}
Although CP-KAN was primarily designed for regression tasks, our experiments on MNIST and CIFAR-10 (detailed in \Cref{sec:appendix_benchmarks}) indicated scaling properties. The model's accuracy improved with network width ((\Cref{fig:mnist_size_vs_acc}, \Cref{fig:cifar10_size_vs_acc})) when QUBO picked lower degrees (\Cref{fig:cifar10_size_vs_acc}, \Cref{fig:mnist_size_vs_acc}) polynomials. Further the QUBO pre-computation time scaled near-linearly ($O(N)$) as shown in \Cref{fig:QUBO_log_scaling}. These findings suggest that CP-KAN might potentially overcome its limitations for classification tasks without prohibitive computational overhead. However, CP-KAN's reliance on smooth polynomials appears to be a challenge for classification tasks requiring sharp decision boundaries, as seen in the Forest Covertype dataset (configuration in \Cref{sec:appendix_covertype_exp_config}) where tree-based methods significantly outperformed it (\Cref{sec:appendix_covertype}). This limitation aligns with theoretical expectations that polynomial bases struggle to approximate the necessary discontinuities effectively. While CP-KAN may require increased width for competitive classification performance, it may achieve this scaling more efficiently than traditional architectures due to its polynomial basis and adaptive degree selection. Rather than uniformly scaling all parameters, CP-KAN's QUBO optimization allows selective complexity increases where needed, offering a nuanced approach to scaling in neural network design.

\section{Conclusion}\label{sec:conclusion}
This work presented CP-KAN, a cumulative Kolmogorov-Arnold polynomial network, illustrating how the classical polynomial approximation theory could be combined with neural architectures through adaptive degree selection. By reformulating the choice of polynomial degree as a QUBO optimization problem solved via simulated annealing, we observed competitive performance in regression tasks, particularly for financial time-series prediction. Our empirical results indicate that CP-KAN achieves competitive performance with considerably fewer parameters compared to the tested traditional deep learning architectures. Although challenges remain, notably the classification performance gap and computational overhead for very large networks, our results suggest that polynomial-based neural architectures could be a viable direction for certain real-world regression tasks.

The CP-KAN framework opens several research directions. First, the polynomial basis offers opportunities for efficient online updates, where coefficients could be adapted as new financial data arrives while maintaining theoretical guarantees. Second, the complementary strengths of polynomials (smooth function approximation) and decision trees (discrete boundaries) suggest hybrid models combining CP-KAN's regression capabilities with tree-based mechanisms, potentially addressing the classification limitations. Additionally, while our current approach selects degrees once, an adaptive system could evolve degrees during training, leading to architectures that adjust complexity to local feature importance~\citep{e24030348}. The connection between QUBO optimization and polynomial degree selection also suggests avenues for further theoretical study regarding optimized polynomial bases. Finally, advanced optimization techniques such as quantum annealing could offer a framework for optimizing network parameters \citep{hauke2020perspectives}, with the adiabatic theorem suggesting potential pathways for exploring the optimization landscape \citep{albash2018adiabatic}. CP-KAN represents an architectural approach toward potentially more principled and efficient neural networks. As machine learning evolves, approaches that blend classical mathematical insights with modern optimization techniques may prove valuable, demonstrating how revisiting fundamental approximation theory through the lens of contemporary machine learning might yield practical benefits and theoretical understanding.

\vspace{0.5cm}
\paragraph*{Acknowledgments.} This work is supported by the National Research Foundation, Singapore, and A*STAR under its CQT Bridging Grant and its Quantum Engineering Programme under grant NRF2021-QEP2-02-P05.

\bibliographystyle{plainnat}
\bibliography{bibliography}

\appendix
\appendix

\noindent
\textbf{Computational Environment:} 
All experiments reported in these appendices were performed on an Apple MacBook Pro (Late 2021) with an M1 Pro chip and 16GB of RAM. With the exception of the LSTM, GRU, Transformer experiments ran on a single H100 GPU.

\section*{Appendix}

\section{Optimization Methods}\label{sec:appendix_optim_formulations}
This section details the mathematical formulation of the different polynomial degree selection methods implemented for the CP-KAN architecture. For a given KAN layer with \( N_{\text{in}} \) inputs and \( N_{\text{out}} \) neurons, the goal is to select a degree \( d_i \in \{0, 1, \ldots, D_{\text{max}}\} \) for each neuron \( i = 0, \ldots, N_{\text{out}}-1 \). Let \( \mathbf{X} \in \mathbb{R}^{B \times N_{\text{in}}} \) be the input data batch and \( \mathbf{Y} \in \mathbb{R}^{B \times N_{\text{out}}} \) be the target data for this layer (either the final targets or dimension-aligned intermediate targets). Let \( \mathbf{y}_i \in \mathbb{R}^{B} \) be the target vector for the \( i \)-th neuron. Let \( T_d(\alpha) \) denote the Chebyshev polynomial of the first kind of degree \( d \), evaluated at \( \alpha \). For the \( i \)-th neuron, the projection is \( \alpha_i = \mathbf{w}_i\mathbf{X} + b_i \). The cumulative transform up to degree \( d \) produces a matrix \( \mathbf{\Phi}_{i,d} \in \mathbb{R}^{B \times (d+1)} \) where the \( k \)-th column is \( T_k(\alpha_i) \). The least-squares coefficients for fitting \( \mathbf{y}_i \) with degree \( d \) are: 
\[
\mathbf{c}_{i,d} = (\mathbf{\Phi}_{i,d}^\top \mathbf{\Phi}_{i,d})^{-1} \mathbf{\Phi}_{i,d}^\top \mathbf{y}_i,
\]
and the resulting prediction is: 
\[
\hat{\mathbf{y}}_{i,d} = \mathbf{\Phi}_{i,d} \mathbf{c}_{i,d}.
\]
The MSE for neuron \( i \) with degree \( d \) is defined as:
\[
\text{MSE}_{i,d} = \frac{1}{B} \| \mathbf{y}_i - \hat{\mathbf{y}}_{i,d} \|^2_2.
\]

\section{Quadratic Unconstrained Binary Optimization (QUBO) Formulation for Degree Selection}\label{sec:appendix_qubo_details}
We start by selecting the optimal polynomial degree for each neuron to minimize the total mean squared error while enforcing that exactly one degree is chosen per neuron. We use binary variables $x_{i,d} \in \{0, 1\}$, where $x_{i,d} = 1$ if neuron $i$ uses degree $d$, and 0 otherwise. These values can be summarized in a Boolean matrix $X$. This problem can be expressed as:
    \begin{align}
    \min_{x_{i,d}\in \{0,1\}} & \sum_{i=1}^N \sum_{d=0}^D c_{i,d}(X)x_{i,d}, \\
    \text{subject to:} & \sum_{d=0}^D x_{i,d} = 1 \quad \forall i, \\
    & x_{i,d} \in \{0,1\} \quad \forall i,d
    \end{align}
    where $c_{i,d}$ represents the MSE cost when neuron $i$ uses degree $d$, $N$ is the number of neurons, and $D$ is the maximum degree allowed.
    Using binary variables $q_{(i,d)}$ (equivalent to $x_{i,d}$), we eliminate explicit constraints by incorporating them as penalty terms:
    \begin{align}
    \min_{\{q_{(i,d)}\}\in\{0,1\}} 
    \;\;\Bigl(\underbrace{\sum_{i,d} \text{MSE}_{i,d} \cdot q_{(i,d)}}_{\text{fidelity cost}}
    \;+\; 
    \underbrace{\alpha\sum_i(\sum_d q_{(i,d)}-1)^2}_{\text{one-hot penalty}}\Bigr)
    \end{align}
    
The problem is solved using a QUBO solver (e.g., simulated annealing) to find the binary assignment $\{q_{(i,d)}\}$ that minimizes the objective. The degree $d$ for which $q_{(i,d)}=1$ is selected for neuron $i$, and the corresponding pre-computed coefficients $\mathbf{c}_{i,d}$ are assigned. This QUBO formulation enables efficient degree selection while maintaining compatibility with both classical simulated annealing and quantum annealing hardware.

\subsection{Integer Programming (IP)}\label{sec:appendix_ip}
The integer programming formulation minimizes the MSE for each neuron independently using a mixed integer linear programming (MILP) solver. For each neuron \( i \), binary variables \( x_{i,d} \in \{0, 1\} \), where \( x_{i,d} = 1 \) if neuron \( i \) uses degree \( d \), and 0 otherwise. The  Objective function (for each neuron \( i \)) is :
    \[
    \text{Minimize} \quad \sum_{d=0}^{D_{\text{max}}} \text{MSE}_{i,d} \cdot x_{i,d}
    \]
    with constraint (for each neuron \( i \)):
    \[
    \sum_{d=0}^{D_{\text{max}}} x_{i,d} = 1
    \]
The MILP problem is solved independently for each neuron \( i \) using a solver like OR-Tools SCIP. The degree \( d \) for which \( x_{i,d}=1 \) is selected, and the corresponding coefficients \( \mathbf{c}_{i,d} \) are assigned.

\subsection{Evolutionary Algorithm (EA)}\label{sec:appendix_ea}
The Evolutionary Algorithm leverages principles of natural selection to efficiently determine an optimal combination of polynomial degrees for all neurons within a given layer simultaneously. Each individual within the algorithm's population is represented by a vector of degrees \( \mathbf{d} = [d_0, d_1, \ldots, d_{N_{\text{out}}-1}] \), where each degree \( d_i \) can take integer values from \( 0 \) up to a predefined maximum degree \( D_{\text{max}} \). The fitness of each individual \( \mathbf{d} \) is calculated as the sum of the mean squared errors for each neuron when using the degrees specified by that individual:
\[
\text{Fitness}(\mathbf{d}) = \sum_{i=0}^{N_{\text{out}}-1} \text{MSE}_{i, d_i}.
\]

The process begins with the initialization of a random population of individuals. For a fixed number of generations, the algorithm performs the following steps: First, it evaluates the fitness of each individual. Then, parents are selected based on their fitness scores, typically favoring those with lower fitness values. Offspring are generated using genetic operators such as crossover, where segments of two parent individuals' degree vectors are combined (for instance, through single-point crossover), and mutation, which involves randomly altering a degree value with a small probability. Optionally, elitism may be applied by preserving one or more of the top-performing individuals from the current generation into the next, ensuring that the best solutions found are not lost. The newly formed offspring then replace the previous population, constituting the new generation. Finally, after completing the predetermined number of generations, the algorithm selects the individual with the lowest fitness as the optimal solution. The degrees from this best individual are then assigned to their respective neurons along with the corresponding pre-computed coefficients \( \mathbf{c}_{i, d_i} \).

\subsection{Greedy Heuristic}\label{sec:appendix_greedy}
The Greedy Heuristic optimization strategy determines the polynomial degree for each neuron sequentially, layer by layer. For a given neuron $i$ in a layer, the algorithm iterates through possible degrees $d = 0, 1, \ldots, D_{\text{max}}$. For each candidate degree $d$, it computes the Chebyshev basis functions $\mathbf{\Phi}_{i,d}$ based on the layer's input data and calculates the optimal least-squares coefficients $\mathbf{c}_{i,d}$ to predict the neuron's target $\mathbf{y}_i$ (either the final task target for the output layer or a dimension-aligned intermediate target for hidden layers, as described in \Cref{sec:appendix_optim_formulations}). The corresponding Mean Squared Error, $\text{MSE}_{i,d}$, is calculated.

The algorithm tracks the degree $d^*$ that yields the lowest MSE encountered so far. A key feature is an early stopping mechanism: the iteration over degrees for neuron $i$ terminates prematurely if the relative improvement in MSE from degree $d-1$ to $d$, defined as $(\text{MSE}_{i,d-1} - \text{MSE}_{i,d}) / \text{MSE}_{i,d-1}$, falls below a predefined \texttt{improvement\_threshold} (e.g., $1\%$). Once the iteration stops (either by reaching $D_{\text{max}}$ or triggering early stopping), the neuron $i$ is assigned the best degree $d^*$ found and its corresponding coefficients $\mathbf{c}_{i,d^*}$. The process then repeats for the next neuron in the layer. After all neurons in the current layer are optimized, their outputs are computed and passed as input to the subsequent layer. The configuration option \texttt{skip\_qubo\_for\_hidden} (though named historically) can bypass this optimization for hidden layers, setting them to a default degree instead.


\section{CP-KAN Training Algorithm Details}\label{sec:appendix_training_algorithm}

This appendix provides the detailed pseudocode for the two-phase training strategy discussed in \Cref{sec:architecture}.

\begin{algorithm}[H]
\caption{CP-KAN Training Algorithm}\label{alg:cpkan-training} 
\small
\begin{algorithmic}[1]
\Require Train/val data $(X_{train}, y_{train})$, $(X_{val}, y_{val})$, network shape, max degree $D$, QUBO threshold $T$, learning rate $\eta$, epochs $E$
\Ensure Trained CP-KAN model

\State Initialize CP-KAN model $\mathcal{M}$ with random weights and biases

\Statex \textbf{Phase 1: Degree Optimization via QUBO}
\For{layer $l = 1$ to $L$}
    \State Set all neurons to degree 3 if $N_l \times (D+1) > T$ \Comment{Skip large layers}
    \If{$N_l \times (D+1) \leq T$}
        \State Compute MSE costs $c_{i,d}$ for each neuron-degree pair (\Cref{sec:appendix_qubo_details})
        \State Construct QUBO matrix $Q$ with costs and one-hot penalties
        \State $\mathbf{q}^* \gets$ Solve QUBO using simulated annealing
        \State Set layer $l$ degrees according to solution $\mathbf{q}^*$
    \EndIf
\EndFor

\Statex \textbf{Phase 2: Parameter Optimization}
\State Collect trainable params $\Theta = \{(\mathbf{w}, b, \mathbf{c})\}$ for all neurons
\State Initialize Adam optimizer with learning rate $\eta$
\For{epoch $e = 1$ to $E$}
    \State Compute loss $\mathcal{L} \gets \text{Loss}(\mathcal{M}(X_{train}), y_{train})$ \Comment{e.g., MSE or CrossEntropy}
    \State Update parameters $\Theta$ using Adam
    \State Monitor validation loss for early stopping (if applicable)
\EndFor
\State \Return Trained model $\mathcal{M}$
\end{algorithmic}
\end{algorithm}


\section{Data Preprocessing and Configuration}\label{sec:appendix_data_config}
\subsection{Jane Street Dataset Configuration}
\label{sec:appendix_jane_street_config}
For the Jane Street Market Prediction dataset, preprocessing involved selecting $79$ numeric features (feature $00$ through feature $78$). Samples were weighted according to a specified column, aligning with the competition's weighted evaluation metrics (weighted MSE and weighted R²). The dataset also included date-based information (date id) to facilitate temporal analysis or validation splits. Experiments typically employed a train-validation split of $70\%$ training and $30\%$ validation data. To manage computational resources effectively, a subset of $200,000$ rows was commonly used.

\begin{table}[ht]
\centering
\caption{Jane Street Dataset Configuration}\label{tab:jane_street_data_config}
\begin{tabular}{@{}ll@{}}
\toprule
Parameter     & Value \\ \midrule
Data Path     & (Kaggle Jane Street Market Prediction dataset path) \\
Rows Used     & 200,000 \\
Train Ratio   & 0.7 \\
Feature Columns & feature\_00 through feature\_78 (79 features) \\
Target Column & 'responder\_6' \\
Weight Column & 'weight' \\
Date Column   & 'date\_id' \\ \bottomrule
\end{tabular}
\end{table}
\subsection{House Sales Dataset Configuration}\label{sec:appendix_house_sales_config}
For the House Sales regression dataset, data was loaded from ``inria-soda/tabular-benchmark'' (reg\_num/house\_sales.csv). The target variable (housing prices) was log-transformed using \texttt{np.log1p()} to reduce skewness and improve model training stability. Features were normalized using \texttt{StandardScaler()} to ensure zero mean and unit variance. For experimentation, $80\%$ of data was used for training and $20\%$ for validation, with a fixed random seed of $42$.

\subsection{Covertype Dataset Configuration}\label{sec:appendix_covertype_config}
For the Covertype classification dataset, data was loaded from ``inria-soda/tabular-benchmark'' (clf\_num/covertype.csv). The class labels were extracted from the ``target'' column (or the last column if named differently). A train-validation split of $80\%-20\%$ was implemented with a fixed random seed of $42$.

\subsection{MNIST and CIFAR-10 Configurations}\label{sec:appendix_image_config}
For the image classification datasets:
\begin{itemize}
    \item \textbf{MNIST:} Standard normalization with mean $0.1307$ and std $0.3081$ was applied. Images were flattened from $28×28$ to $784$-dimensional input vectors.
    \item \textbf{CIFAR-10:} Standard normalization with mean ($0.5$, $0.5$, $0.5$) and std ($0.5$, $0.5$, $0.5$) was applied. Images were flattened from $3$×$32$×$32$ to $3072$-dimensional input vectors.
    \item \textbf{Batch Size:} $64$ was used as the standard batch size for both datasets.
    \item \textbf{Training:} Models used CrossEntropy loss and targets were converted to one-hot format for QUBO optimization.
\end{itemize}


\section{Hyperparameter Configurations for Jane Street Experiments}\label{sec:appendix_jane_street_hyperparams}
\subsection{Experiment 1: CP-KAN Optimization Method Comparison}\label{sec:appendix_js_exp1}
This experiment compared the performance of different degree optimization strategies within the CP-KAN framework. A grid search was performed across various architectural and training hyperparameters, applying the same grid to each optimization method tested.

\begin{table}[ht]
\centering
\captionof{table}{Hyperparameter Grid Search for CP-KAN Optimization Methods (Jane Street)}\label{tab:jane_street_optim_hyperparams}
\begin{tabular}{@{}ll@{}}
\toprule
Hyperparameter             & Values Searched                                           \\ \midrule
Optimization Methods Tested & QUBO, IntegerProgramming, Evolutionary, GreedyHeuristic \\ \midrule
\multicolumn{2}{@{}l}{\textbf{Grid Search Parameters (Applied to all methods):}}      \\
Max Degree                 & \{3, 5, 7, 9\}                                            \\
Hidden Size                & \{16, 20, 24, 32\}                                        \\
Hidden Degree              & \{3, 5, 7\}                                               \\
Learning Rate              & \{1e-2, 5e-3, 1e-3\}                                      \\ \bottomrule
\end{tabular}
\end{table}

\subsection{Experiment 2: CP-KAN vs MLP Architecture Comparison}\label{sec:appendix_js_exp2}

This experiment performed a grid search to compare the performance of the CP-KAN architecture against various MLP architectures (see \Cref{tab:jane_street_kan_mlp_hyperparams}). The goal was to find suitable base configurations for further analysis, such as the stability tests in Experiment 3.

\begin{table}[ht]
\centering
\caption{Hyperparameter Grid Search for KAN vs MLP (Jane Street – Pre‑Degradation)}
\label{tab:jane_street_kan_mlp_hyperparams}
\begin{tabular}{@{}lll@{}}
\toprule
Model       & Hyperparameter  & Values Searched           \\ \midrule
CP-KAN      & Max Degree      & \{5, 7, 9\}               \\
            & Hidden Size     & \{16, 20, 24, 28\}        \\
            & Hidden Degree   & \{3, 5, 7\}               \\
            & Learning Rate   & \{0.01, 0.005\}           \\ \midrule
MLP         & Hidden Size     & \{20, 24, 28\}            \\
            & Depth           & \{2, 3, 4\}               \\
            & Dropout         & \{0.1, 0.15\}             \\
            & Learning Rate   & \{0.01, 0.005\}           \\ \bottomrule
\end{tabular}
\end{table}

\subsection{Experiment 3: Training Stability / Degradation Analysis}\label{sec:appendix_js_exp3}
This experiment investigated the training stability and performance degradation using the best performing CP-KAN and MLP configurations identified from Experiment 2 (\Cref{tab:jane_street_kan_mlp_hyperparams}). These chosen architectures (CP-KAN with 20 hidden neurons; MLP with hidden size 24 and depth 3) were then trained with different learning rates to observe stability patterns. The specific fixed parameters for these chosen models are detailed in \Cref{tab:jane_street_degradation_config}.

\begin{table}[ht]
\centering
\caption{Fixed Configurations (Best from Exp.~2) for Degradation Analysis (Jane Street)}
\label{tab:jane_street_degradation_config}
\begin{tabular}{@{}lll@{}}
\toprule
Model    & Parameter               & Fixed Value                \\ \midrule
CP-KAN   & Network Shape           & {[}Input Dim, \textbf{20}, 1{]}  \\
         & Max Degree              & 5                          \\
         & Complexity Weight       & 0.0                        \\
         & Trainable Coefficients  & True                       \\
         & Skip QUBO for Hidden    & False                      \\
         & Default Hidden Degree   & 5                          \\ \midrule
MLP      & Hidden Size             & \textbf{24}                \\
         & Depth                   & \textbf{3}                 \\
         & Dropout Rate            & 0.1 (default in build\_mlp) \\ \midrule
\multicolumn{3}{@{}l}{\textbf{Variable Parameter for this Experiment:}} \\
KAN \& MLP & Learning Rate         & \{1e-2, 1e-3, 1e-4, 1e-5\} \\ 
\bottomrule
\end{tabular}
\end{table}

\subsection{Experiment 4: Hyperparameter Grids for KAN Architecture Comparison (Jane Street)}\label{sec:appendix_js_exp4}

This experiment compared various KAN architectures against each other and other models like MLPs on the Jane Street dataset. The following hyperparameter grids were searched for each model type. Note that CP-KAN variants shared a similar grid structure. The base \texttt{learning\_rate} for training loops was 1e-3, but could be overridden by specific model configurations if included in their grid search.

\begin{table}[ht]
\centering
\caption{Hyperparameter Grid Search for KAN Architecture Comparison}
\label{tab:kan_arch_comparison_hyperparams}
\begin{tabular}{@{}lll@{}}
\toprule
Model Type          & Hyperparameter           & Values Searched                      \\ \midrule
CP-KAN (All variants) & Max Degree               & \{5, 7, 9\}                          \\
                    & Complexity Weight        & \{0.05, 0.1, 0.2\}                   \\
                    & Trainable Coefficients   & \{False, True\}                      \\
                    & Skip QUBO for Hidden     & \{False, True\}                      \\
                    & Default Hidden Degree    & \{3, 4, 5\}                          \\
                    & Hidden Dims              & [20] (Fixed)                     \\ \midrule
Original-KAN (Spline) & Hidden Dim             & \{16, 20, 24, 32\}                   \\
                    & Num (Spline Intervals)   & \{3, 5, 7\}                          \\
                    & K (Spline Order)         & \{2, 3, 4\}                          \\ \midrule
Wavelet-KAN         & Hidden Dim              & \{16, 20, 24, 32\}                   \\
                    & With Batch Norm         & \{True, False\}                      \\
                    & Wavelet Type            & 'mexican\_hat' (Fixed)               \\ \midrule
Jacobi-KAN          & Hidden Dim              & \{16, 20, 24, 32\}                   \\
                    & Degree                  & \{3, 5, 7\}                          \\
                    & Parameter \texttt{a}    & \{0.5, 1.0, 1.5\}                    \\
                    & Parameter \texttt{b}    & \{0.5, 1.0, 1.5\}                    \\ \midrule
Chebyshev-KAN       & Hidden Dim              & \{16, 20, 24, 32\}                   \\
(ChebyKAN)          & Degree                  & \{3, 5, 7, 9\}                       \\ \midrule
Taylor-KAN          & Hidden Dim              & \{16, 20, 24, 32\}                   \\
                    & Order                   & \{2, 3, 4, 5\}                       \\
                    & Add Bias                & True (Fixed)                         \\ \midrule
RBF-KAN             & Hidden Dim              & \{16, 20, 24, 32\}                   \\
                    & Num Centers             & \{5, 10, 15, 20\}                    \\
                    & Alpha (Shape Param)     & \{0.5, 1.0, 2.0\}                    \\ \midrule
MLP                 & Hidden Dims             & [32], [64], [128], [32, 32], [64, 64] \\
                    & Dropout                & \{0.1, 0.2, 0.3\}                    \\
                    & Use Batch Norm         & \{True, False\}                      \\ \bottomrule
\end{tabular}
\end{table}

\section{Hyperparameter Grid Search for House Sales Dataset}\label{sec:appendix_house_sales_hyperparams}
This experiment performed a hyperparameter grid search comparing CP-KAN and MLP architectures on the House Sales regression dataset (using log-transformed target values).

\begin{table}[ht]
\centering
\captionof{table}{Hyperparameter Grid Search for House Sales Dataset}\label{tab:house_sales_hyperparams}
\begin{tabular}{@{}lll@{}}
\toprule
Model       & Hyperparameter           & Values Searched                      \\ \midrule
CP-KAN      & Max Degree               & \{5, 7, 9, 11\}                      \\
            & Hidden Size (\texttt{network\_shape}) & \{20, 32, 64\}                       \\
            & Default Hidden Degree    & \{3, 5, 7, 9\}                       \\
            & Complexity Weight        & \{0.0, 0.001, 0.01\}                 \\
            & Learning Rate            & \{1e-4, 5e-4, 1e-3\}                 \\ \midrule
MLP         & Depth                    & \{2, 3, 4, 5\}                       \\
            & Hidden Size              & \{32, 64, 128\}                      \\
            & Dropout Rate             & \{0.1, 0.2\}                         \\
            & Learning Rate            & \{1e-4, 1e-3\}                       \\
            & Weight Decay             & \{0.0, 0.001\}                       \\
            & Batch Size               & \{64, 128, 256\}                     \\ \bottomrule
\end{tabular}
\end{table}


\section{Optimization Method Evaluation (MNIST, CIFAR-10)}\label{sec:appendix_optim_eval_mnist_cifar}

This experiment aimed to compare different CP-KAN optimization methods (QUBO, IntegerProgramming, Evolutionary, GreedyHeuristic) using a hyperparameter grid search. The target datasets were MNIST and CIFAR-10, though the execution was adapted based on initial results on MNIST. On MNIST, the IntegerProgramming, Evolutionary, and GreedyHeuristic methods yielded near-random accuracy (\(\approx 11.35\%\)), while QUBO achieved significantly better performance (37.48\% accuracy). Due to the poor results from the other methods on MNIST, the planned grid search was \textbf{not} conducted for them on CIFAR-10.

Where applicable, base training parameters such as \texttt{num\_epochs} were set to $50$. The hyperparameter grid search was applied as follows:
\begin{itemize}
    \item \textbf{MNIST:} All four methods (QUBO, IP, Evo, Greedy) were evaluated using the full grid. Only QUBO demonstrated promising results.
    \item \textbf{CIFAR-10:} Only QUBO was evaluated using the grid search. A few runs with IP configurations were attempted, but yielded worse-than-random performance (\(\approx 9\%\) accuracy).
\end{itemize}

\begin{table}[ht]
\centering
\captionof{table}{Hyperparameter Grid Search Definition for Optimization Methods}\label{tab:optim_method_grid_search_actual}
\begin{tabular}{@{}lll@{}}
\toprule
Method(s) Applied To        & Hyperparameter           & Values Defined in Grid         \\ \midrule
QUBO (MNIST, CIFAR-10)      & Max Degree               & \{3, 5, 7\}                    \\
IP, Evo, Greedy (MNIST only)& Hidden Layers            & [64], [128], [32, 32]      \\
                            & Default Hidden Degree    & \{3\} (Fixed)                  \\
                            & Complexity Weight        & \{0.01\} (Fixed)               \\
                            & Trainable Coefficients   & \{True\} (Fixed)               \\
                            & Learning Rate            & \{0.001, 0.0003, 0.0001\}      \\ \midrule
Evolutionary (MNIST only)   & Population Size          & \{20, 50\}                     \\
                            & Generations              & \{30\} (Fixed)                 \\ \bottomrule
\end{tabular}
\end{table}


\section{Covertype Classification Configuration}\label{sec:appendix_covertype_exp_config}

This experiment compared CP-KAN against an MLP on the Covertype classification dataset (\texttt{test\_tabular\_data.py}). The KAN and MLP architectures were automatically configured to target a similar parameter count (\texttt{desired\_params = $10000$}), rather than using a grid search. Standard RandomForest and XGBoost models were also trained as baselines using fixed configurations.

\begin{table}[ht]
\centering
\captionof{table}{Fixed Training Parameters for KAN/MLP on Covertype}\label{tab:covertype_fixed_params}
\begin{tabular}{@{}ll@{}}
\toprule
Parameter     & Fixed Value \\ \midrule
Epochs        & 50 (or 500, check script execution) \\
Batch Size    & 4096        \\
Learning Rate & 1e-3        \\
Optimizer     & Adam        \\
Loss Function & CrossEntropy \\ \bottomrule
\end{tabular}
\end{table}

\begin{table}[ht]
\captionof{table}{KAN/MLP Configurations Tested on Covertype (Target Params \(\approx\) 10k)}\label{tab:covertype_configs}
\begin{tabular}{@{}lll@{}}
\toprule
Model    & Parameter               & Fixed Value                    \\ \midrule
CP-KAN   & Network Shape           & Determined by calculating summing all the trainable parameters \\
         & Max Degree              & 10                             \\
         & Complexity Weight       & 0.0                            \\
         & Trainable Coefficients  & False                          \\
         & Skip QUBO for Hidden    & False                          \\
         & Default Hidden Degree   & 2                              \\ \midrule
MLP      & Architecture            & Determined by \texttt{approximate\_mlp}      \\
         & Hidden Activation       & ReLU                           \\
         & Layers include          & BatchNorm1d, Dropout           \\ \bottomrule
\end{tabular}
\end{table}

\begin{table}[ht]
\centering
\captionof{table}{Baseline Model Configurations for Covertype}\label{tab:covertype_baseline_configs}
\begin{tabular}{@{}lll@{}}
\toprule
Model          & Parameter           & Fixed Value           \\ \midrule
RandomForest   & n\_estimators       & 100                   \\
               & random\_state       & 42                    \\
               & Other               & scikit-learn defaults \\ \midrule
XGBoost        & n\_estimators       & 100                   \\
               & random\_state       & 42                    \\
               & use\_label\_encoder & False                 \\
               & eval\_metric        & logloss             \\
               & Other               & XGBoost defaults      \\ \bottomrule
\end{tabular}
\end{table}


\section{Scaling and Classification Benchmarks}\label{sec:appendix_benchmarks}

\subsection{MNIST Benchmark}\label{sec:appendix_mnist}
\begin{figure}[t]
  \centering
  \begin{subfigure}[t]{0.49\linewidth}
    \centering
    \includegraphics[width=\textwidth]{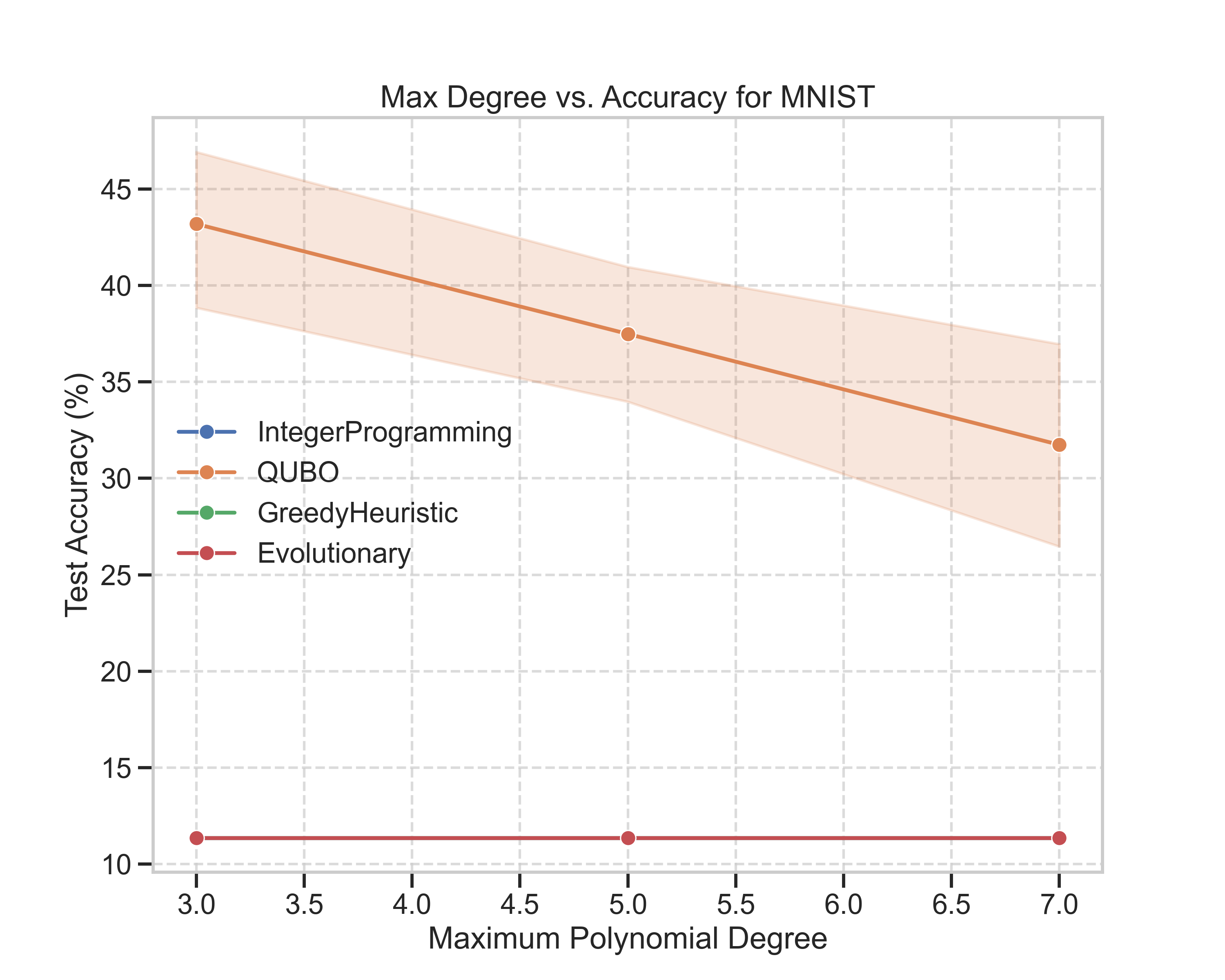}
    \caption{Test accuracy vs.\ maximum polynomial degree $D$.}
    \label{fig:mnist_degree_vs_acc}
  \end{subfigure}
  \hfill
  \begin{subfigure}[t]{0.49\linewidth}
    \centering
    \includegraphics[width=\textwidth]{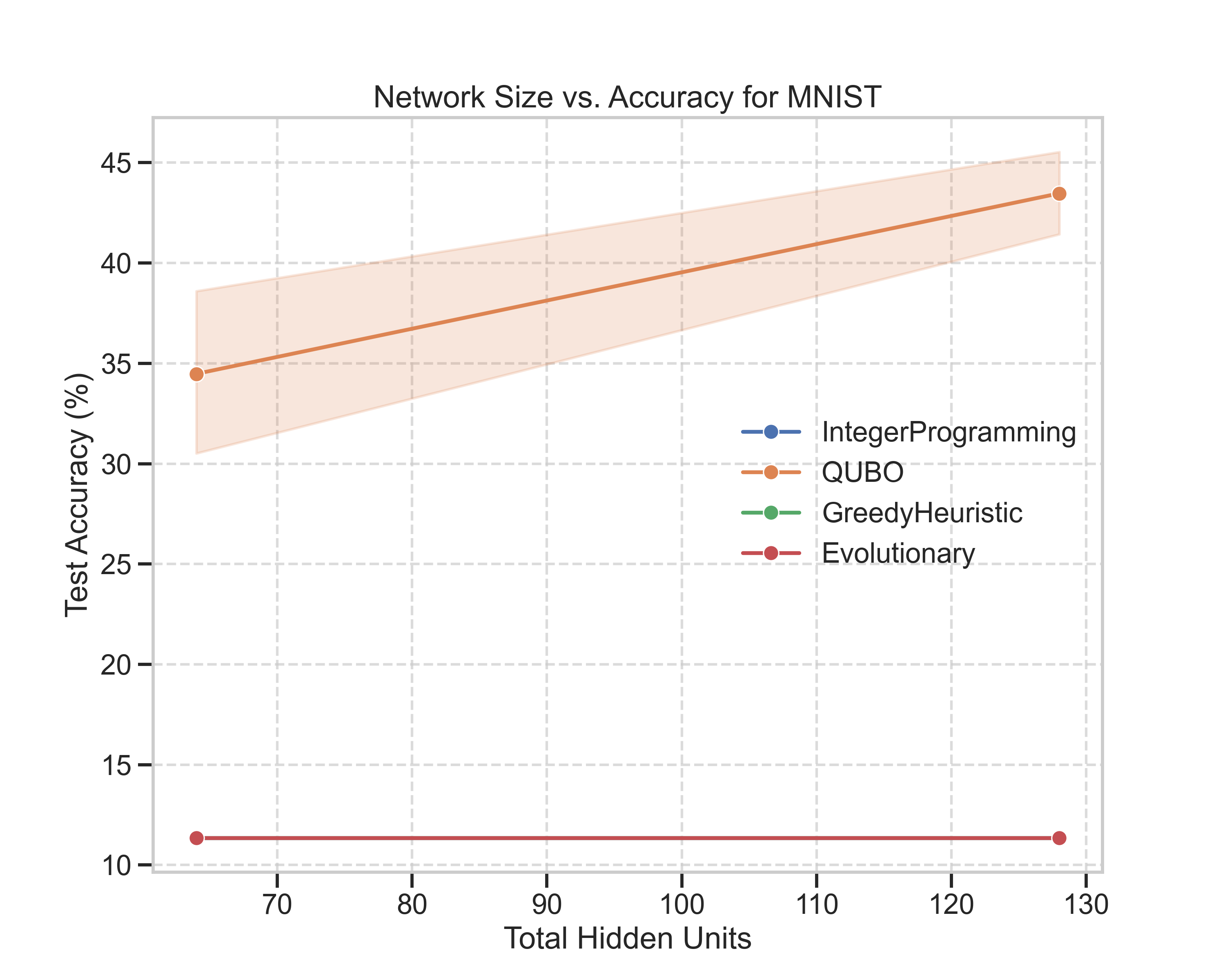}
    \caption{Test accuracy vs.\ total hidden units.}
    \label{fig:mnist_size_vs_acc}
  \end{subfigure}
  \caption{Impact of model complexity on MNIST test accuracy for CP-KAN solved via QUBO (dashed) and integer programming (solid).}
  \label{fig:mnist_complexity_vs_acc}
\end{figure}

\begin{figure}[ht]
   \centering
    \includegraphics[width=0.99\textwidth]{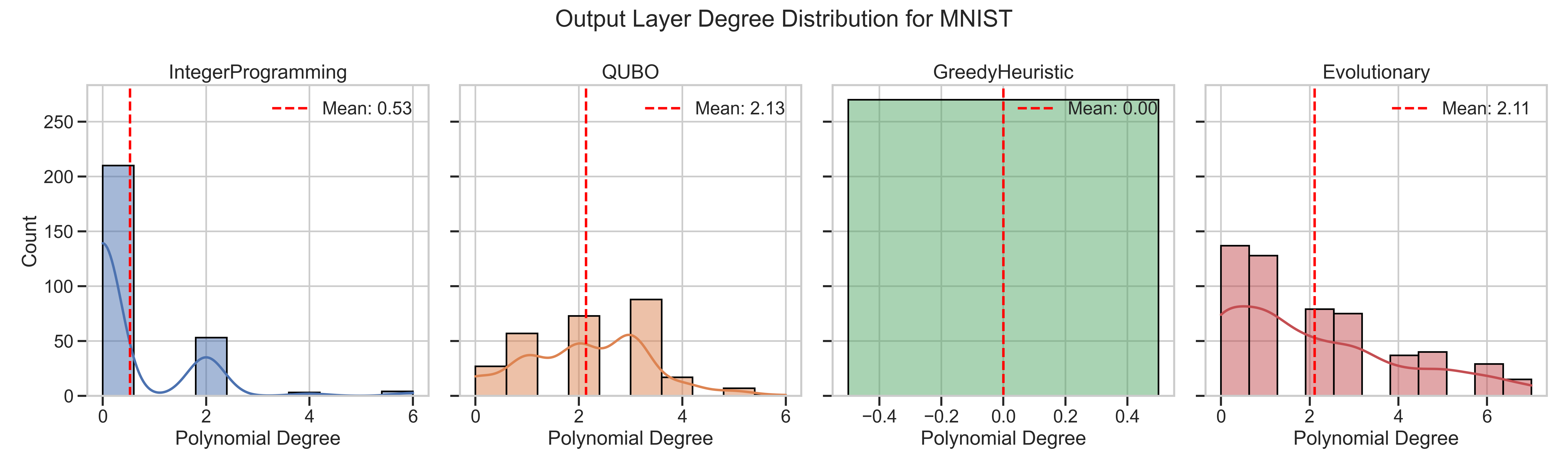}
    \caption{Comparison of polynomial degree distributions selected by different optimization methods for the output layer neurons on the MNIST dataset. This visualization provides context on optimizer behavior and preference for specific degrees.}\label{fig:mnist_degree_dist_appendix}
\end{figure}

To provide further insight into the scaling properties of CP-KAN and the comparison between optimization methods, we present additional results from experiments conducted on the MNIST dataset. These results parallel the analysis performed on CIFAR10 in the main text. \Cref{fig:mnist_degree_vs_acc} shows the impact of varying the maximum allowed polynomial degree, while \Cref{fig:mnist_size_vs_acc} illustrates the effect of network size on test accuracy. For context, \Cref{fig:mnist_degree_dist_appendix} shows the distribution of degrees selected by different optimizers in one configuration.

\subsection{CIFAR-10 Benchmark}\label{sec:appendix_cifar10}
\begin{figure}[!ht]
  \centering
  \begin{subfigure}[t]{0.49\textwidth}
    \centering
    \includegraphics[width=\linewidth]{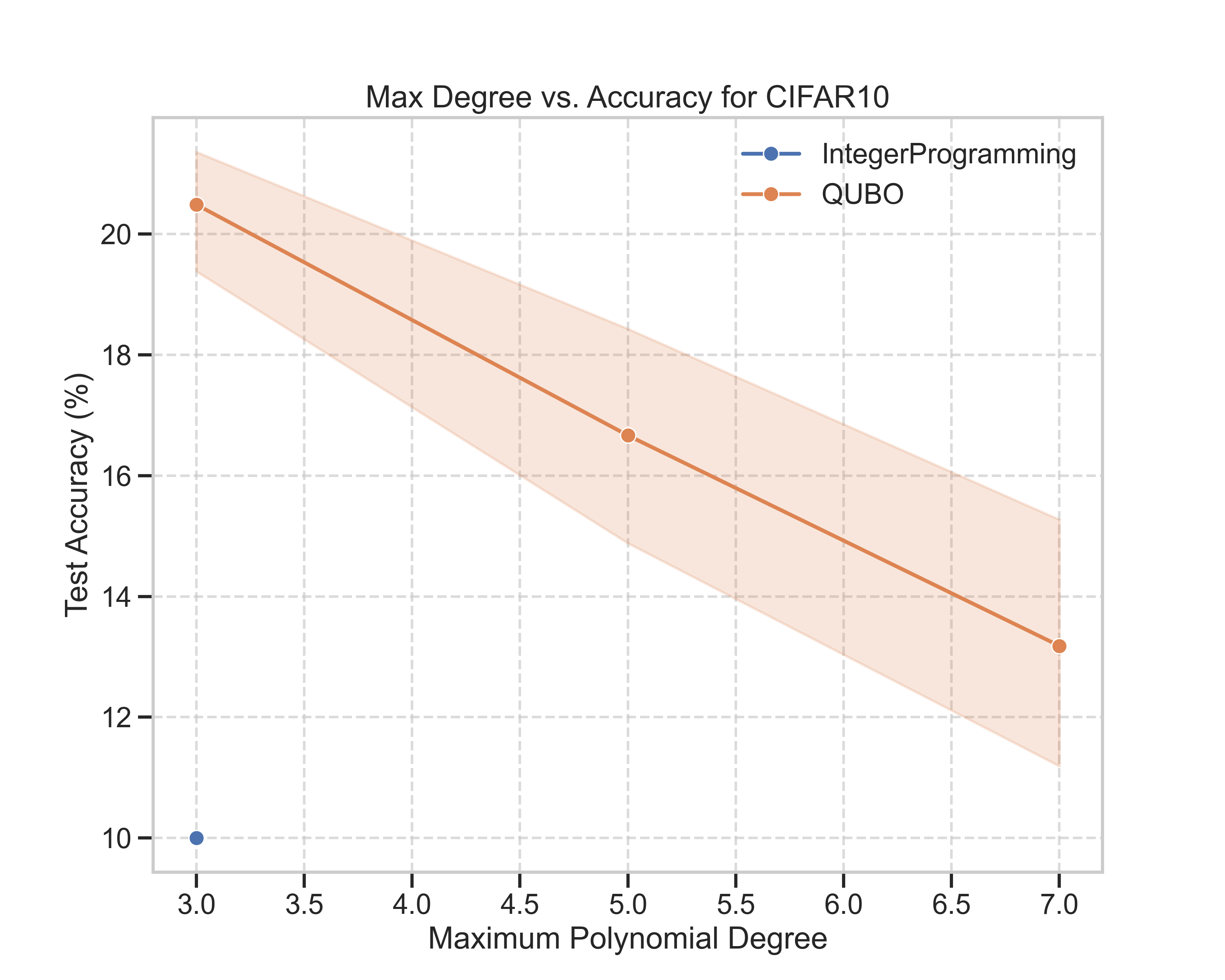}
    \caption{Test accuracy vs.\ maximum polynomial degree $D$.}
    \label{fig:cifar10_degree_vs_acc}
  \end{subfigure}
  \hfill
  \begin{subfigure}[t]{0.49\textwidth}
    \centering
    \includegraphics[width=\linewidth]{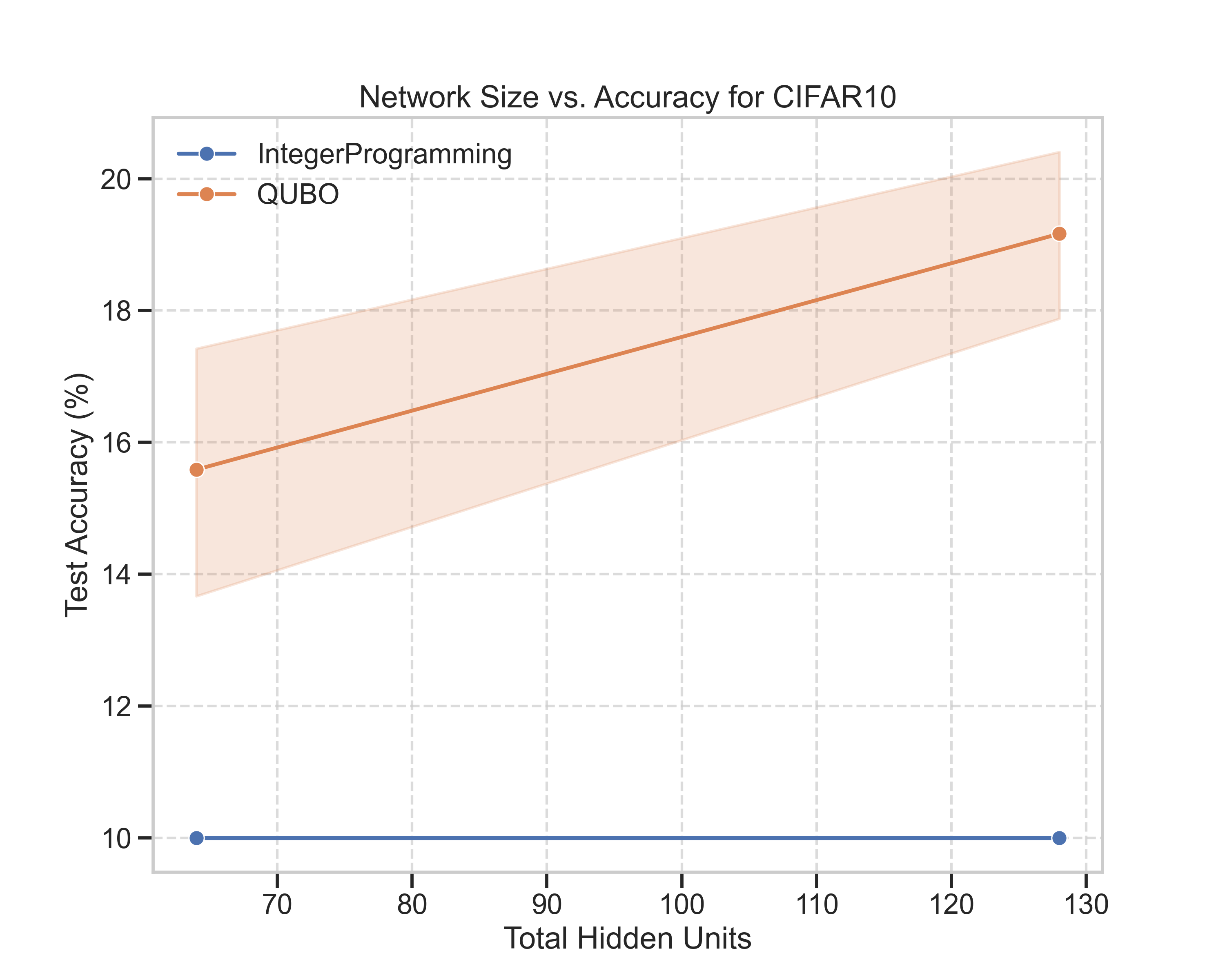}
    \caption{Test accuracy vs.\ total hidden units.}
    \label{fig:cifar10_size_vs_acc}
  \end{subfigure}
  \caption{Impact of model complexity on CIFAR-10 test accuracy for CP-KAN optimized with QUBO.}
  \label{fig:cifar10_complexity_vs_acc}
\end{figure}

\justify First, we present the computational scaling results mentioned in the main text.
\begin{figure}[ht]
    \centering
    \includegraphics[width=0.8\columnwidth]{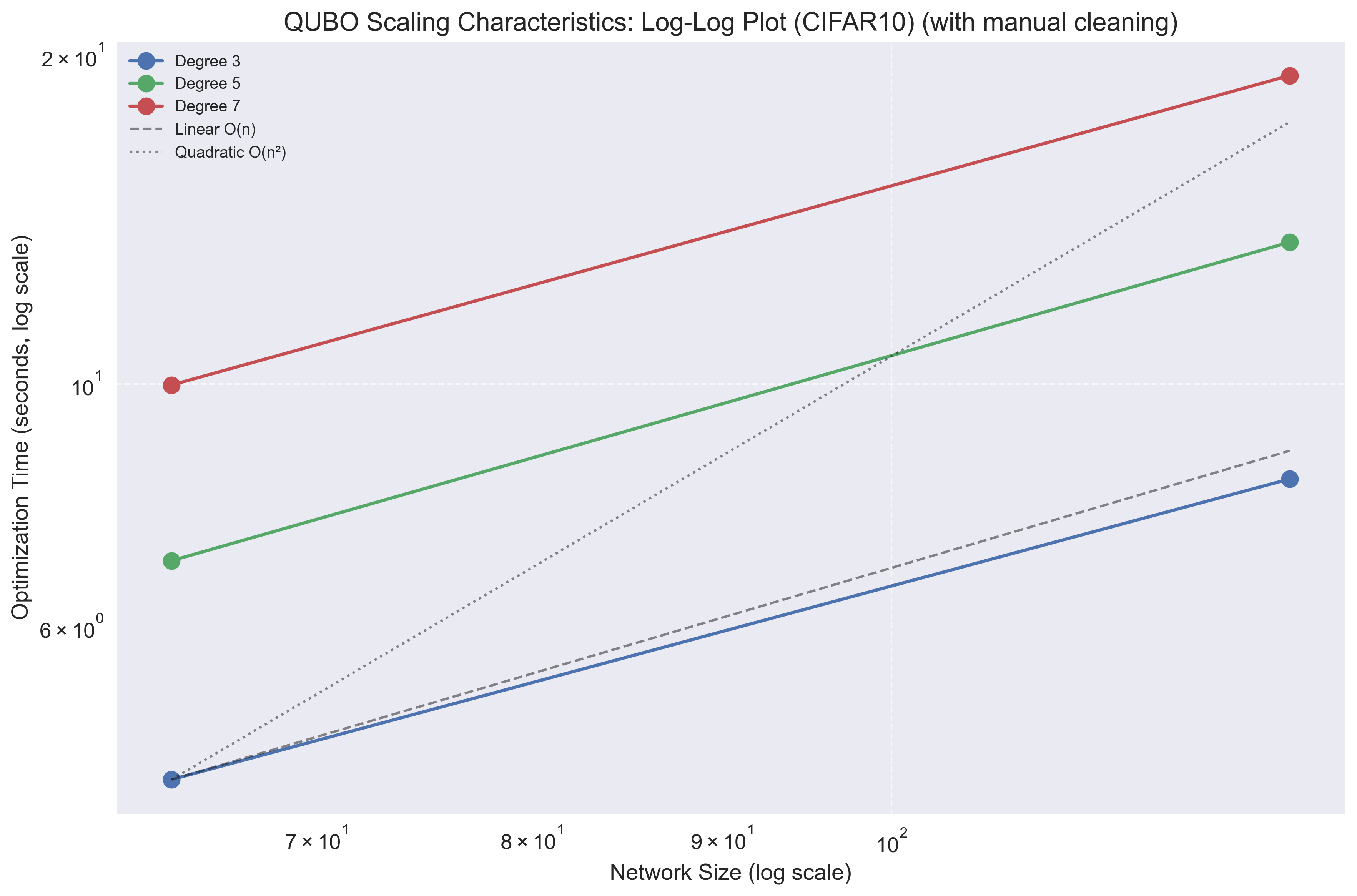}
    \caption{Log-log plot illustrating the computational scaling of the QUBO optimization time versus network size on CIFAR10. Results are shown for different maximum degrees ($D=3,5,7$). The scaling closely follows the linear $O(n)$ reference line, indicating linear complexity. Similar results were observed in the MNIST experiments.}\label{fig:QUBO_log_scaling}
\end{figure}
Similar experiments were conducted on the CIFAR-10 dataset to assess scaling behavior in a more complex image classification task. \Cref{fig:cifar10_degree_vs_acc} shows the relationship between maximum polynomial degree and accuracy, while \Cref{fig:cifar10_size_vs_acc} illustrates the effect of network size.

The CIFAR-10 results corroborate the findings from MNIST: QUBO optimization facilitates positive scaling with network size, whereas excessively high polynomial degrees can hinder performance. The QUBO solver often selects lower degrees, suggesting the importance of adaptable complexity.

\subsection{Forest Covertype Classification}\label{sec:appendix_covertype}
We also evaluated CP-KAN on the Forest Covertype dataset, a standard tabular classification benchmark known for requiring models to capture potentially sharp decision boundaries. As detailed in Appendix~\ref{sec:appendix_covertype_exp_config}, CP-KAN (optimized with QUBO) was compared against an MLP configured for a similar parameter count (\(\approx\)10k) and standard baseline models (RandomForest, XGBoost).
The results indicated that while CP-KAN achieved reasonable accuracy, it was outperformed by both the MLP and significantly by the tree-based methods. This aligns with the expectation that CP-KAN\'s polynomial basis functions are inherently better suited for approximating smooth, continuous functions (as in regression) rather than the potentially discontinuous decision boundaries required for optimal performance on Covertype. Tree-based models excel in such scenarios by recursively partitioning the feature space. This experiment highlights a limitation of the current CP-KAN architecture for certain types of classification problems, reinforcing its primary strengths in regression and smooth function modeling.

\subsection{QUBO Solver Timings}\label{sec:appendix_qubo_timings}
The QUBO optimization step (Phase 1), typically solved via simulated annealing, has a runtime primarily influenced by the number of binary variables in the QUBO matrix (number of neurons $N \times (\text{max degree } D+1)$) and the time required to calculate the MSE cost matrix (which depends on the batch size $B$ and input data size). \Cref{fig:QUBO_log_scaling} illustrates the near-linear scaling complexity observed with respect to network size $N$. To provide concrete examples, \Cref{tab:qubo_timings_specific} details the execution time for this QUBO step on the hardware specified in Appendix A.

\begin{table}[!ht]
\centering
\captionof{table}{Execution time (\textit{mean}~$\pm$~\textit{std.~dev.} over $n$ runs) for the QUBO optimization step (Phase 1, simulated annealing) on an Apple MacBook Pro (Late 2021, M1 Pro chip) for different CP-KAN layer sizes ($N$ neurons) and maximum Chebyshev degrees ($D$). Batch size used for MSE calculation was $64$.}\label{tab:qubo_timings_specific}
\begin{tabular}{@{}llll@{}}
\toprule
Layer Size (N) & Degree 3 (D=3) & Degree 5 (D=5) & Degree 7 (D=7) \\
\midrule
64  & 4.35s $\pm$ 0.10 (n=5) & 6.89s $\pm$ 0.17 (n=6) & 9.98s $\pm$ 0.95 (n=6) \\
128 & 8.19s $\pm$ 0.02 (n=3) & 13.49s $\pm$ 0.29 (n=3) & 19.17s $\pm$ 0.32 (n=3) \\
\bottomrule
\end{tabular}
\end{table}

\section{Recurrent \& Transformer Models (Embedding Trick \& Architecture Details)}\label{sec:appendix_recurrent_transformer}

This appendix provides a reproducible account of how we apply LSTM, GRU, and Transformer models to tabular data using an embedding trick that treats each feature as a ``time step'' (for LSTM/GRU) or a ``token'' (for Transformer). We also describe a larger, competition-style GRU variant that was tested.

\subsection{Feature-as-Sequence Embedding Trick}\label{sec:appendix_embedding_trick}
For tabular data with $D$ features, standard sequence models expect input shape $\bigl[B, \text{seq\_len}, \text{input\_dim}\bigr]$. To adapt tabular data $\mathbf{X} \in \mathbb{R}^{B \times D}$, we first reshape it to $\mathbf{X}' = \mathbf{X}.\texttt{unsqueeze}(-1) \in \mathbb{R}^{B \times D \times 1}$, treating each feature as a time step or token. Then, we apply a linear layer (e.g., \texttt{nn.Linear(1, embed\_dim)}) to project this into a higher embedding dimension $d_{\text{embed}}$, resulting in $\mathbf{X}'' \in \mathbb{R}^{B \times D \times d_{\text{embed}}}$. This embedded sequence $\mathbf{X}''$ is then fed into the LSTM, GRU, or Transformer model. Finally, the model's output (e.g., the last hidden state for RNNs or pooled tokens for Transformers) is mapped to the target dimension via a final linear layer.

\subsection{LSTM and GRU Models}\label{sec:appendix_lstm_gru_merged}
We used standard \texttt{torch.nn.LSTM} or \texttt{torch.nn.GRU} with \texttt{batch\_first=True}. Key hyperparameters include hidden dimension $d_{\text{hidden}}$ and number of layers $N_{\mathrm{layers}}$. The final hidden state $\mathbf{h}_{\text{final}} \in \mathbb{R}^{B \times d_{\text{hidden}}}$ is passed to a linear layer. We also tested a ``competition-style'' GRU variant with multiple layers (e.g., $\{500\}$ or $\{250, 150, 150\}$), inter-layer dropout, and a post-GRU feed-forward stack (e.g., $[500, 300]$), inspired by Kaggle solutions but adapted for typical batch sizes.

\subsection{Transformer Model (Encoder Only)}\label{sec:appendix_transformer_condensed}
Our Transformer encoder approach used \texttt{torch.nn.TransformerEncoderLayer} (with model dimension $d_{\mathrm{model}}$, number of heads $n_{\mathrm{heads}}$, feedforward dimension $d_{\mathrm{ffn}}$) stacked into a \texttt{torch.nn.TransformerEncoder} with $N_{\mathrm{layers}}$. After embedding features to $[B, D, d_{\mathrm{model}}]$, the encoder outputs shape $[B, D, d_{\mathrm{model}}]$. We typically average pool across the $D$ tokens before a final linear projection.

\subsection{Weighted Loss and Metric}\label{sec:appendix_loss}
During training, we minimize a weighted MSE (or WMSE) using sample weights $w_i$, and validate using the competition's weighted $R^2$ metric, employing early stopping based on validation $R^2$.

\subsection{Implementation Notes}\label{sec:appendix_impl_notes_condensed}
Models were implemented in PyTorch. The embedding trick involves \texttt{x.unsqueeze(-1)} followed by \texttt{nn.Linear}. Standard library modules were used for LSTM, GRU, and Transformer components. Training used the AdamW optimizer, logging the best validation $R^2_{\text{weighted}}$. This feature-as-sequence approach allows applying sequence models to tabular data but requires careful tuning.

\section{Theoretical Details for Mean Reversion}\label{sec:appendix_mean_reversion_theory}
This appendix provides the formal statement of the generator decomposition theorem and the error bound related to the Ornstein-Uhlenbeck process discussed in \Cref{sec:discussion}.

The efficiency of Chebyshev polynomials stems from their approximation properties for smooth functions. Assuming a function $g(x)$ defined on $[-1,1]$ has $k$ continuous derivatives, its Chebyshev expansion $g(x) \approx \sum_{i=0}^\infty c_i T_i(x)$ exhibits coefficients $c_i$ that decay algebraically, typically as $O(i^{-(k+1)})$ \citep{decayRateChebyshev}. This rapid decay ensures that smooth functions, like the time-varying transfer function $A_t(\omega)$ in locally stationary processes (after appropriate rescaling of $\omega$), can be accurately approximated using a finite number of Chebyshev terms, underpinning their suitability for CP-KAN.

\begin{theorem}[Generator Decomposition \citep{infinitesimalGen2007}]
For the mean-reverting SDE $dX_t = \theta(\mu - X_t)\,dt + \sigma\,dW_t,$ 
the infinitesimal generator $\mathcal{A}$ admits a Chebyshev expansion: 
\begin{align}
    \mathcal{A}f(x) 
    &= \theta(\mu - x)f'(x) + \tfrac12\sigma^2f''(x) \notag \\
    &= \sum_{n=0}^\infty \lambda_n \langle f, T_n\rangle T_n(x),
\end{align}
with $|\lambda_n|\le C(1+n^2).$
\end{theorem}

This decomposition provides rigorous justification for the use of Chebyshev polynomials in modeling mean-reverting processes often found in finance \citep{stockVol2016}. Furthermore, for the Ornstein-Uhlenbeck process, tight error bounds can be established for approximations like those potentially learned by CP-KAN \citep{spectralPerturb2018}:
\begin{align}
    \mathbb{E}[\|X_t - \hat{X}_t\|^2] 
    \;\le\;
    C_1 e^{-\theta t}\|X_0-\mu\|^2 \;+\; C_2\frac{\log d}{d}. 
\end{align}
Here, $\hat{X}_t$ represents the approximation using polynomials up to degree $d$. The bound highlights the interplay between the exponential decay due to mean reversion ($heta$) and the approximation error related to the polynomial degree $d$.

\end{document}